\documentclass{article}

\usepackage{microtype}
\usepackage{graphicx}
\usepackage{subcaption}
\usepackage{booktabs} 

\usepackage{hyperref}
\usepackage{xcolor}

\usepackage[accepted]{icml2024}

\usepackage{amsmath}
\usepackage{amssymb}
\usepackage{mathtools}
\usepackage{amsthm}

\usepackage[capitalize,noabbrev]{cleveref}

\theoremstyle{plain}

\theoremstyle{definition}

\theoremstyle{remark}

\usepackage{cuted}

\usepackage[textsize=tiny]{todonotes}

\usepackage{bm}
\usepackage{tabularx}
\newcolumntype{Y}{>{\centering\arraybackslash}X}

\usepackage{xurl}
\usepackage{placeins}

\icmltitlerunning{Dirac--Bianconi Graph Neural Networks - Enabling long-range graph predictions}

\usepackage{amsmath,amsfonts,bm}

\def\eqref#1{equation~\ref{#1}}

\def\1{\bm{1}}

\def\mB{{\bm{B}}}

\def\mL{{\bm{L}}}

\def\mW{{\bm{W}}}

\DeclareMathAlphabet{\mathsfit}{\encodingdefault}{\sfdefault}{m}{sl}
\SetMathAlphabet{\mathsfit}{bold}{\encodingdefault}{\sfdefault}{bx}{n}

\def\gE{{\mathcal{E}}}

\def\gG{{\mathcal{G}}}

\def\gN{{\mathcal{N}}}

\def\sC{{\mathbb{C}}}

\def\sR{{\mathbb{R}}}

\begin{document}

\twocolumn[



\icmltitle{Dirac--Bianconi Graph Neural Networks -- Enabling Non-Diffusive Long-Range Graph Predictions}


\icmlsetsymbol{equal}{*}


\begin{icmlauthorlist}
\icmlauthor{Christian Nauck}{pik}
\icmlauthor{Rohan Gorantla}{ue,ue2}
\icmlauthor{Michael Lindner}{pik}
\icmlauthor{Konstantin Schürholt}{usg}
\icmlauthor{Antonia S. J. S. Mey}{ue}
\icmlauthor{Frank Hellmann}{pik}
\end{icmlauthorlist}

\icmlaffiliation{pik}{Potsdam Institute for Climate Impact Research, Telegrafenberg A31, 14473 Potsdam, Germany}
\icmlaffiliation{ue}{EaStCHEM School of Chemistry, University of Edinburgh, EH9 3FJ, UK}
\icmlaffiliation{usg}{AIML Lab, University of St. Gallen, Rosenbergstrasse 30, CH-9000 St. Gallen, Switzerland}
\icmlaffiliation{ue2}{School of Informatics, University of Edinburgh, EH8 9AB, UK}


\icmlcorrespondingauthor{Christian Nauck}{nauck@pik-potsdam.de}
\icmlcorrespondingauthor{Frank Hellmann}{hellmann@pik-potsdam.de}

\icmlkeywords{Machine Learning, ICML}

\vskip 0.3in
\editorsListText
\vskip 0.3in
]



\printAffiliationsAndNotice{}  

\begin{abstract}

The geometry of a graph is encoded in dynamical processes on the graph. Many graph neural network (GNN) architectures are inspired by such dynamical systems, typically based on the graph Laplacian. Here, we introduce Dirac--Bianconi GNNs (DBGNNs), which are based on the topological Dirac equation recently proposed by Bianconi. 
Based on the graph Laplacian, we demonstrate that DBGNNs explore the geometry of the graph in a fundamentally different way than conventional message passing neural networks (MPNNs). 
While regular MPNNs propagate features diffusively, analogous to the heat equation, DBGNNs allow for coherent long-range propagation.
Experimental results showcase the superior performance of DBGNNs over existing conventional MPNNs for long-range predictions of power grid stability and peptide properties. This study highlights the effectiveness of DBGNNs in capturing intricate graph dynamics, providing notable advancements in GNN architectures.

\end{abstract}

\section{Introduction}

Understanding the geometric and physical properties of real-world data is crucial for AI applications in physics, chemistry, and robotics. For the common case of graph structured data, various graph neural network (GNN) architectures have been developed \cite{wuComprehensiveSurveyGraph2021}. 

GNNs have been motivated in various ways. Spectral approaches see the eigenbasis of the graph Laplacian $\mL$ as playing a similar role as that of Fourier modes in spatial data.

Spatial convolution approaches, instead, build on the idea of aggregating features by considering spatial neighborhoods. Message passing is one of the most prominent implementations of this idea. For the edge connecting nodes $i$ and $j$, we construct a message $e_{ij}$ from inputs $x^\text{in}_i$, $x^\text{in}_j$ at its ends, and some invariant edge features $e^f_{ij}$. Then each node aggregates messages on the edges connected to it to produce an output. Denoting general, potentially non-linear, functions by $f$, $g$, $h$, and for the set of neighbors $\mathcal{N}_i$ of node $i$, the most general structure usually assumed for message passing neural networks (MPNNs) is:
\begin{align}
\label{eq:messagepassing}
    e_{ij} &= g\left(x^\text{in}_i, x^\text{in}_j, e^f_{ij}\right),\\
    x^\text{out}_i &= f\left(x^\text{in}_i, \sum_{j \in \mathcal{N}_i} h(e_{ij}, x^\text{in}_j)\right).
\end{align}
Here, $g$ can compute simple differences across edges and attention scores. There is a large body of work exploring the potential of this style of architecture, including variations of updates to edge features. In this context, the case where edge features are updated simultaneously with node features from layer to layer (see for example \citet{yangNENNIncorporateNode2020,chenEdgeFeaturedGraphAttention2021,heGeneralizationViTMLPMixer2023,zhouCoembeddingEdgesNodes2023}) has not been explored as extensively:
\begin{align}
\label{eq:edge-feature-update}
    e^\text{out}_{ij} &= g\left(x^\text{in}_i, x^\text{in}_j, e^\text{in}_{ij}\right),\\
    x^\text{out}_i &= f\left(x^\text{in}_i, \sum_{j \in \mathcal{N}_i} h(e^\text{in}_{ij}, x^\text{in}_j)\right).
\end{align}

In contrast to ordinary MPNNs, the variable $e$ cannot be eliminated from the equations. Edge features propagate from layer to layer. Note that this also falls outside the edge update model of \citet{battaglia2018relational}, which considers node updates of the form $\sum h(e^\text{out}_{ij}, x^\text{in}_j)$.

\looseness-1
%

\vspace{-1pt}
\textbf{Exploring GNNs as dynamical systems} These structures resemble that of a discrete dynamical system on a network. This perspective has led to novel approaches to GNN architectures that draw inspiration from dynamical systems to overcome limitations observed in earlier architectures. A notable example is the work of \citet{ruschGraphCoupledOscillatorNetworks2022}, who introduce neural networks based on inertial Kuramoto oscillator networks \citep{kuramotoSelfentrainmentPopulationCoupled1975, acebronKuramotoModelSimple2005, rodriguesKuramotoModelComplex2016}. They argue that oversmoothing -- the observation that many GNN architectures tend to average out features across the graph when iterated too deeply -- is analogous to synchronization in such dynamical systems. The fact that the synchronous manifold is unstable for oscillator dynamics under a natural parameter condition then suggests that its architecture should not suffer from oversmoothing. \looseness-1

GNN architectures have been inspired by theoretical physics in numerous related works. Notable works draw from geometric curvature \citep{toppingUnderstandingOversquashingBottlenecks2022}, discrete dynamical systems \citep{oonoGraphNeuralNetworks2021}, ordinary differential equations \citep{poliGraphNeuralOrdinary2021}, as well as partial differential equations and their discretization schemes \citep{chamberlainGRANDGraphNeural2021, chamberlainBeltramiFlowNeural2021,eliasofPDEGCNNovelArchitectures2021}. 

Generally, these works are based on dynamical systems defined in terms of variants of the graph Laplacian. The paradigmatic example of a dynamical system defined by the Laplacian is the heat equation, which describes diffusion of a density $\rho$ in space ($r$) and time ($t$):
\begin{align}
    \partial_t \rho(r, t) = - \Delta \rho(r, t).
\end{align}
The process of heat spreading involves the gradual equalization of temperature throughout the system. This spreading can be adjusted in various ways, leading to the development of analogous GNN architectures. Advection plays a role in moving heat around \citep{eliasofFeatureTransportationImproves2023, zhaoGraphNeuralConvectionDiffusion2023}, different regions may exhibit varying heat conductivities, and anomalous diffusion can impact the speed at which the system reaches equilibrium \citep{maskeyFractionalGraphLaplacian2023}.

The main motivation of our work is to investigate a new class of physical processes that can be used as the basis for GNN architectures. Consider the way that light travels: It can slow down in a medium, reflect, refract, and travel long distances without losing its shape. The underlying equation has oscillating plane waves as solutions, which can be combined into wave packets that maintain their shape while traveling long distances. Fiber optics provide a great example of how a medium can shape the path of a signal while the wave packet keeps its shape, allowing a light pulse to transmit information over hundreds of kilometers. Unfortunately, there is no direct analogy to electromagnetic equations in graphs, since there is no natural gradient operator.

\paragraph{Inspiration from the Dirac equation on networks}
 Instead, we base our work on the topological Dirac equation on networks, recently introduced by \citet{bianconiTopologicalDiracEquation2021}. The Dirac equation is one of the fundamental equations of quantum mechanics. It describes the evolution of the Dirac field, representing most elementary particles such as electrons, protons, and quarks. Like electromagnetic equations, it allows wave packets to maintain their shape.
 
 The Dirac equation is based on the Dirac operator, a square root of the Laplacian. It is obtained by mixing spatial derivatives with transformations in `internal space', i.e., transformations between different components of the Dirac field. \citet{bianconiTopologicalDiracEquation2021} builds on earlier work on quantum information processing by \citet{lloydQuantumAlgorithmsTopological2016}, introducing a Dirac operator on simplicial complexes to create a topological Dirac equation for graphs. \looseness-1

The topological Dirac equation falls outside the dynamics captured by the standard MPNN framework shown in \Cref{eq:messagepassing}, as it treats edges and nodes on the same footing. It is thus an example of the less studied layer of the type described by \Cref{eq:edge-feature-update} with propagating edge features. Given that it is based on a physical equation that allows for wave packets to propagate long distances without losing their shape, we expect this architecture to be of interest where oversmoothing and long distance dependencies are important. Furthermore, the fact that edges and nodes are treated equally is also appropriate for some tasks. \looseness-1

%

\vspace{-1pt}
\textbf{Potential for longe-range predictions}
One motivating application for GNNs with long-range capabilities inspired by dynamical systems is predicting stability properties of power grids. \citet{ringsquandlPowerRelationalInductive2021} show that GNNs with 13 or more layers are needed to achieve good performance in predicting stability properties of power grids with roughly 300 buses. This differs from many commonly used benchmark datasets where GNNs with only 2-3 layers perform best. This suggests that architectures that tend to oversmooth will struggle in the context of power grids. Power grids have node and edge features of similar nature and importance, suggesting the need to treat edges and nodes similarly, rather than considering the former as just a coupling for the latter. In addition, the complex topological aspects of the graph topology play a significant role in shaping the dynamic properties in power grids. A striking example is shown by \citet{nitzbonDecipheringImprintTopology2017}, who find that certain desynchronization modes of their power grid model occur only in specific topological settings, regardless of the node features.


\vspace{-1pt}
\textbf{Our key contributions are:} 
\vspace{-2pt}
\begin{itemize}\vspace{-.3cm}
    \item We analyze the diffusive character of conventional MPNNs by visualizing the trajectories of signals through the GNNs. \looseness-1\vspace{-.25cm}
    \item We generalize the topological Dirac equation and use it to define the novel Dirac--Bianconi T-Step Layer (DBTS).\looseness-1\vspace{-.25cm}
    \item We show experimentally that this layer allows for shape-preserving long-range propagation of feature activation along the graph. MPNN architectures with the same random weights do not exhibit this behavior. We also show empirically that the Dirichlet energy does not go to zero under repeated applications of this layer. 
    \item To validate its performance on benchmark datasets, we use an architecture with several DBTS layers and skip connections. We find that the new layer shows superior performance on challenging power grid tasks with crucial long-range dependencies. On molecular tasks for predicting peptide properties from long-range benchmark datasets, DBGNN outperforms conventional message passing (MPNN) methods while using a quarter of the parameters, and is competitive with transformer-based GNNs.\looseness-1\vspace{-.25cm}
\end{itemize}
Our work presents a key contribution to improving GNNs and providing long-range capabilities by propagating node and edge features. This paves the way for assembling more complex graph-based datasets that rely on edge features and long-range propagation. 
In the following sections, we introduce Bianconi's Dirac operator for graphs, as well as the topological Dirac equation and our generalization thereof in more detail. This provides the basis for the introduction of the DBGNN layer. We then present experimental results and compare the performance with benchmark models. \looseness-1
%
%
%
%
%
\section{Background}
\textbf{Notation:} 
Graphs $\gG$ consist of nodes $\gN$ and edges $\gE$. Each edge occurs twice, with the two possible orientations, and we write an edge $e$ as an ordered pair $[i, j] \in \mathcal{E}$ of nodes $i, j \in \gN$. The set of neighbors of node $i$ is denoted as $\gN_i$. The space of features on an edge/node is called $F_{e/n}$, the space of all edge features of our graph $\gG$ is $F_{e}^{\gE}$, and $F_{n}^{\gN}$ for node features. \looseness-1
%

The structure of a graph can be described using the incidence matrix $\mB \in \sR^{|\gN|\times|\gE|} $:\looseness-1
\begin{align}
B_{ie} = \begin{cases} +1 \text{ if } e = [i, j]\\
 -1 \text{ if } e = [j, i]\\
  0 \text{ otherwise.}
\end{cases}
\end{align}
\textbf{Introducing the Dirac operator}: The incidence matrix $\mB$ maps from the edge space to the node space. It can be checked that matrix $\mB \mB^\dagger$ gives the usual Laplacian matrix, where $^\dagger$ denotes the conjugate transpose. Consider a vector of node features $x$ and a vector of edge features $e$. Bianconi's Dirac operator on graphs is given by:

\begin{align}
\partial_{DB} \begin{pmatrix}x\\ e\end{pmatrix} = \left( \begin{matrix}
0 & b \mB \\
(b \mB)^\dagger & 0
\end{matrix} \right) \begin{pmatrix}x\\ e\end{pmatrix} = \begin{pmatrix}b \mB e\\ b^* \mB^\dagger x\end{pmatrix}
\end{align}

for some $b \in \sC$. We see that this operator maps node features to edge features, and vice versa. The equation considered by \citet{bianconiTopologicalDiracEquation2021} contains a mass term $\beta$. Taking $x$ and $e$ as a function of $t \in \sR{}$, the topological Dirac equation is
\begin{align}
i \partial_t \begin{pmatrix}x(t)\\ e(t)\end{pmatrix} &= \left( \partial_{DB} + \begin{pmatrix}\beta & 0 \\ 0 & - \beta\end{pmatrix}\right) \begin{pmatrix}x(t)\\ e(t)\end{pmatrix}\\
&= \left( \partial_{DB} + \mW_\beta\right) \begin{pmatrix}x(t)\\ e(t)\end{pmatrix}\label{eq:top-DB} \; .
\end{align}

\citet{bianconiTopologicalDiracEquation2021} discusses in detail how this equation relates to the motivating Dirac equation for certain graphs; however, it is not at all obvious what the analog of plane waves and wave packets should be for arbitrary topologies. This is a topic of ongoing research.

At the same time, the spectral properties of this equation make it clear that it is fundamentally different from the heat equation. Recall that the heat equation is $\partial_t x = \bm\Delta x$ with solutions given in terms of eigenvectors $v_i$ and eigenvalues $\lambda_i$ by $\sum_i e^{\lambda_i t} v_i$. Since $\bm\Delta$ is a negative semi-definite matrix, we have $\lambda_i \leq 0$. For connected graphs, only one $\lambda$ is zero and all solutions decay to the corresponding eigenvector. 

In contrast, the operator $\partial_{DB} + \mW_\beta$ has an equal number of positive and negative eigenvalues \citep{bianconiTopologicalDiracEquation2021}. These are bounded away from zero by $|\beta|$. Thus, the evolution according to this operator cannot equilibrate -- there are always expanding and contracting directions. Furthermore, if we note the imaginary unit in front of the time-derivative on the left-hand side, $i \partial_t$, the positive and negative eigenvalues correspond to rotating and counter-rotating oscillations in time, which can be superimposed to approximate temporal envelopes for wave packets. Spatially, these oscillations couple the node and edge spaces together. We therefore hypothesize that such wave packets propagate directly into the graph.

\textbf{Details on the Dirac operator:}
For completeness, the following subsection discusses further aspects and motivation for the Dirac operator, although the details are not necessary for understanding the rest of the paper. \looseness-1

In differential geometry, any square root of the Laplacian (typically on a vector bundle over a Riemannian manifold) is called a Dirac operator. While the eigenvalues of the Laplacian operator show how diffusive processes disperse, they do not distinguish between different directions in the manifold. In contrast, those of the Dirac operators typically do so by coupling directions in space to different directions in the bundle. This can be illustrated by the simplest example. For the tangent bundle over $\sR$, the Dirac operator $- i \partial_x$ has eigenvalues $\pm 1$ for eigenmodes $e^{\pm i t}$. The Laplace operator has the same eigenmodes, but both correspond to the eigenvalue $1$. Exponentiating the Laplacian leads to the heat kernel, which smooths out the differences and leads to equilibration in the long run: All states converge to the kernel of the Laplace operator. In contrast, exponentiating $- i \partial_x$ simply induces shifts along the real axis: $e^{- s i \partial_x}f(x) = f(x - s)$, which can be verified by taking the Fourier transform. \looseness-1

%
As already noted above, the incidence matrix $\mB$ satisfies $\mL = \mB \mB^\dagger$. In the context of homological algebra, $\mB$ is called the boundary operator, and $\mB^\dagger \mB$ is the so-called one-down Laplacian connecting edges to edges. Such boundary operators play a central role in the extension of GNNs to simplicial complexes. A principled approach to simplicial message passing using this approach is given in \citet{bodnarWeisfeilerLehmanGo2021}. \looseness-1

In our context, the relationship between the incidence matrix and the Laplacian suggests that the incidence matrix should play a role similar to that of the Dirac operator. However, since it maps between edges and nodes, it cannot be used directly to update features, which motivates the definition above.

As discussed in detail in \citep{bianconiTopologicalDiracEquation2021}, this Dirac operator does not capture directionality in the same way as the usual Dirac operators of quantum mechanics. For example, on a regular lattice, the operator does not distinguish between edges that are parallel or orthogonal to each other. However, it does mix edge features and node features in non-trivial ways. In addition, its eigenstates encode topological features of the graph on both the edges and the nodes. \looseness-1

Since $\partial_{DB}$ squares the block matrix consisting of the usual graph Laplacian and the one-down Laplacian, iteration of this operator does not introduce any interesting non-diffusive dynamics on the feature space of the graph. However, as the spectral analysis shows, the dynamical equations based on it are genuinely different. We will see further evidence of this difference in the numerical experiments that study the spreading behavior of feature activation.

\section{The Dirac--Bianconi Layer}

\begin{figure*}[htb]
    \centering
    \includegraphics[width=0.4\textwidth]{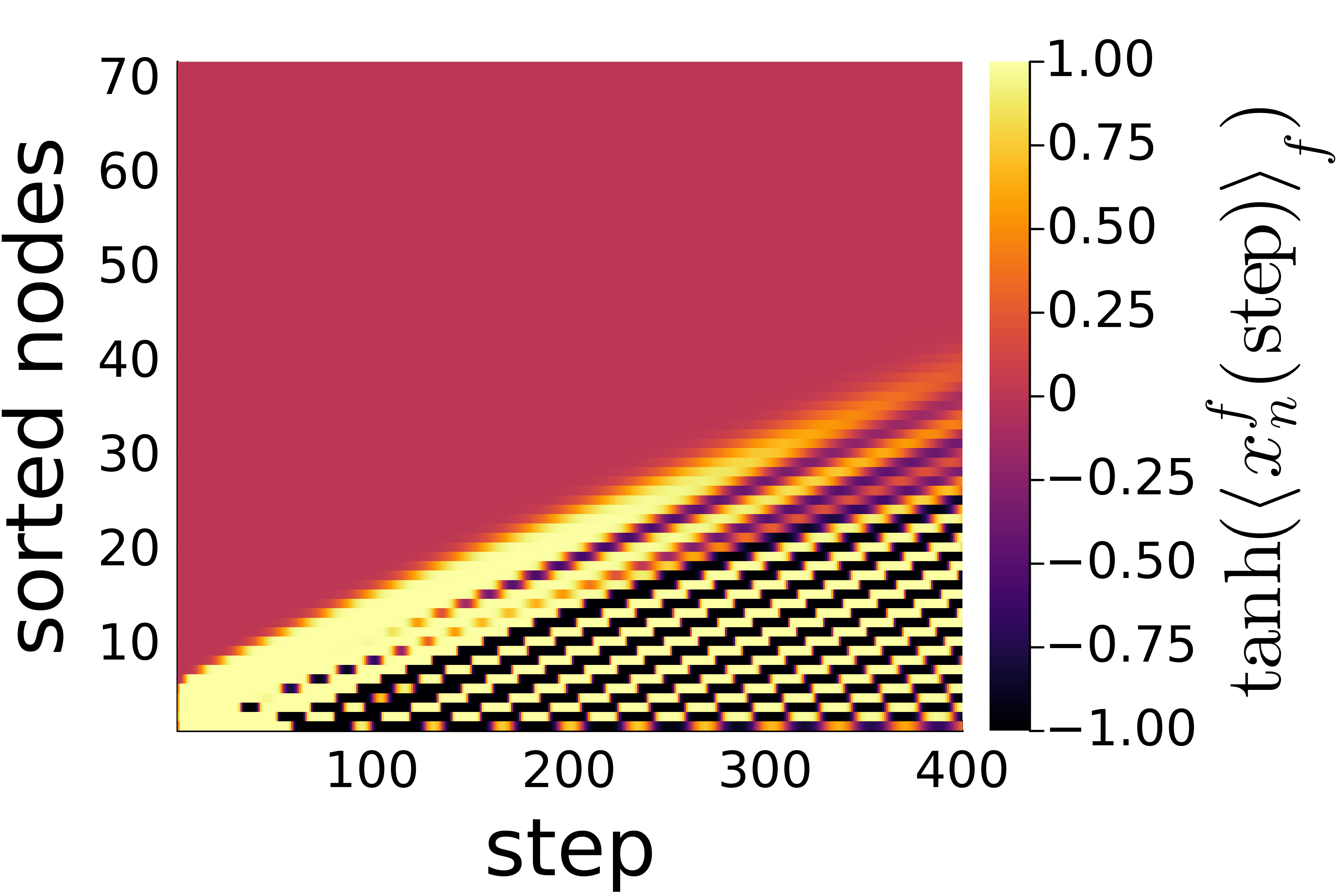}
    \includegraphics[width=0.4\textwidth]{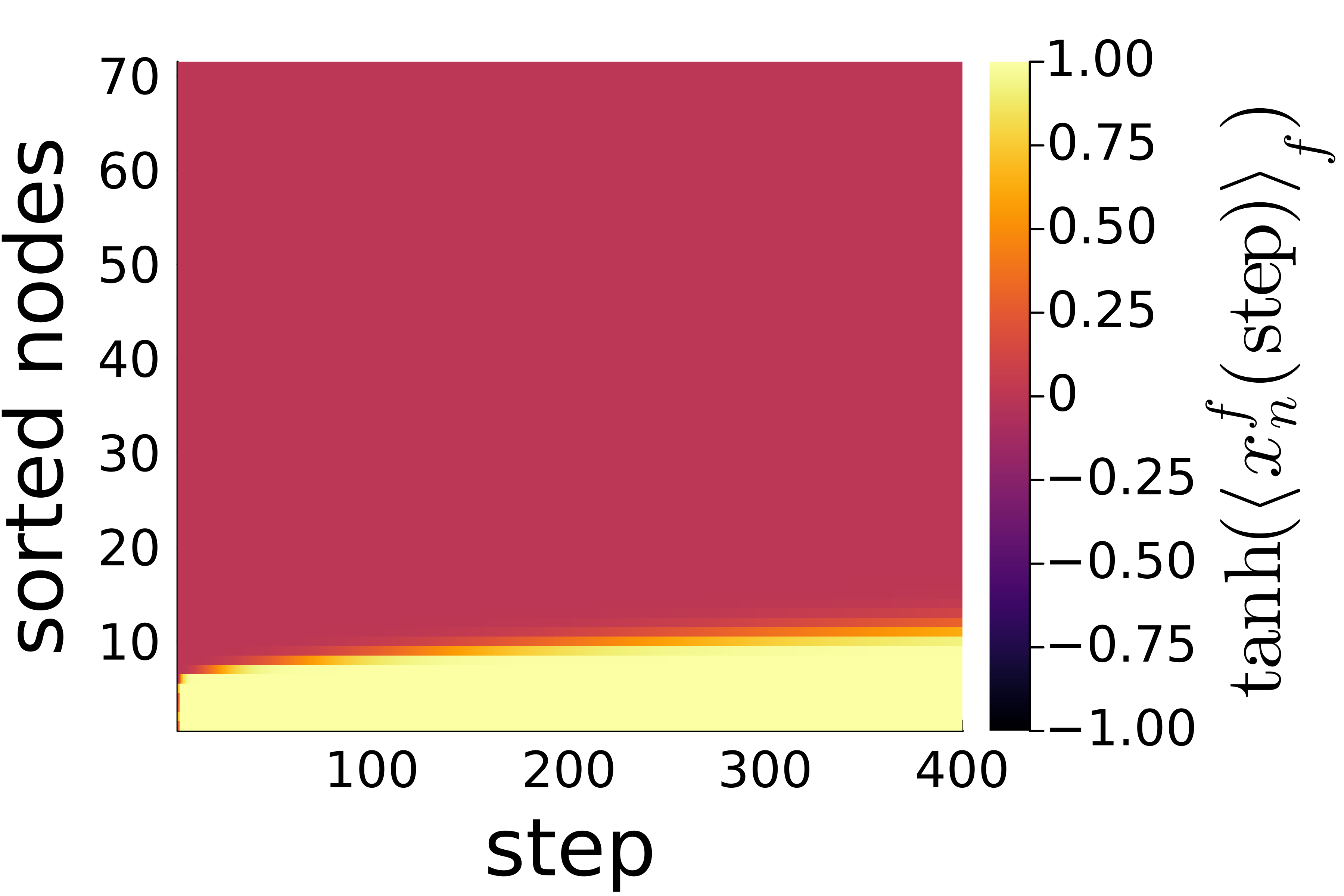}
    \caption{Feature activation versus steps of linear DB \Cref{eq:linDB} (left) and MPNN \Cref{eq:MPNN} (right) on a path graph, $d_n = d_f = 1$, same random weights. The initial condition has all features zero, except at node 1, where the node features are randomly activated. Linear DB shows activation moving linearly down the graph, while MPNN shows diffusion.}
    \vspace{-.3cm}
    \label{fig:spreading}
\end{figure*}

We consider the Euler discretization of \Cref{eq:top-DB} and introduce higher-dimensional feature spaces of dimension $d_n$ for nodes and dimension $d_e$ for edges, $F_n = \sR^{d_n}$ and $F_e = \sR^{d_e}$. Consider coupling matrices $\mW^{ne} \in \sR^{d_n \times d_e}$ and $\mW^{en} \in \sR^{d_e \times d_n}$, and mass matrices $\mW^{n}_\beta \in \sR^{d_n \times d_n}$, $\mW^{e}_\beta \in \sR^{d_e \times d_e}$, we can write
\begin{align}
    x_i(t+1) &= x_i(t) +  \mW^{ne} \sum_{j \in \mathcal{N}_i} e_{ij}(t) + \mW^{n}_\beta  x_i(t), \label{eq:linDB}\\
    e_{ij}(t+1) &= e_{ij}(t) + \mW^{en} (x_i(t) - x_j(t)) - \mW^{e}_\beta  e_{ij}(t). \nonumber
\end{align}
We call this equation the generalized linear Dirac--Bianconi equation. Here $F_n^\gN=\sR^{d_n |\gN|}$ denotes the total node feature space and $F_e^\gE$ the total edge feature space. However, after applying $\partial_{DB}$ we also have the space of one edge feature space per node, $F_e^\gN$, and one node feature space per edge, $F_n^\gE$. The action of $\mW^{ne}$ and $\mW^{en}$ then maps us back to the original $F_n^\gN$ and $F_e^\gE$. \looseness-1
This equation contains both the wave behavior and the expanding/contracting dynamics as a special case. The oscillatory behavior occurs when we mimic the imaginary unit by making the right-hand side of \Cref{eq:linDB} antisymmetric: $\beta_{n/e} = -\beta_{n/e}^\dagger$, and $\mW^{ne} = -{\mW^{en}}^\dagger$.
This should be compared to a simple linear message passing neural network (MPNN)-style dynamics with edge weights, omitting activation functions:
\begin{align}
    x_i(t+1) &= x_i(t) +  \mW_n^\text{MPNN} \sum_{j \in \mathcal{N}_i} e_{ij}(t) + \beta_{n}  x_i(t), \label{eq:MPNN}\\
    e_{ij}(t) &= \mW_e^\text{MPNN} (x_i(t) - x_j(t)) \nonumber
\end{align}
while the edge messages change over time, there is no time evolution of the edge features themselves. The messages can be trivially eliminated, while this is not the case for \Cref{eq:linDB}. This changes the way dynamics spread in the network, see \Cref{fig:spreading}. The propagation of edge features allows oscillations to couple directly to the graph structure, enabling waves to propagate. Note that this is true even if the dataset itself has no edge features. \looseness-1

\Cref{eq:linDB} is of the general form considered in \citet{bodnarWeisfeilerLehmanGo2021}, but the specific form here was not considered there. From the perspective of simplicial topology, the main difference is that we use the boundary and coboundary operators, together with a simultaneous update of nodes and edges, but no Laplacian. For simplicial two-complexes this has been considered \cite{bunchSimplicial2ComplexConvolutional2020}, but we are not aware of any previous work of this kind on graphs. \looseness-1

To conclude this section, Laplacian operators, which are common message passing frameworks, lead to smoothing with loss of information. In contrast, the Dirac operator, with its form of the wave equation, does not lead to oversmoothing, which can also be seen visually in \Cref{fig:spreading}. \looseness-1
\section{Dirac--Bianconi Graph Neural Networks}
In the following, we use the generalized linear Dirac--Bianconi equation (\Cref{eq:linDB}) to define a novel GNN layer for problems where long-range interactions in the graph are expected to play a profound role. This is achieved by explicitly considering node and edge features in a coupled system and by using the Dirac operator, which does not smooth wave signals. \looseness-1

\begin{figure*}[th]
    
     \begin{subfigure}[b]{.25\textwidth}
        \centering
        \includegraphics[width=0.8\columnwidth]{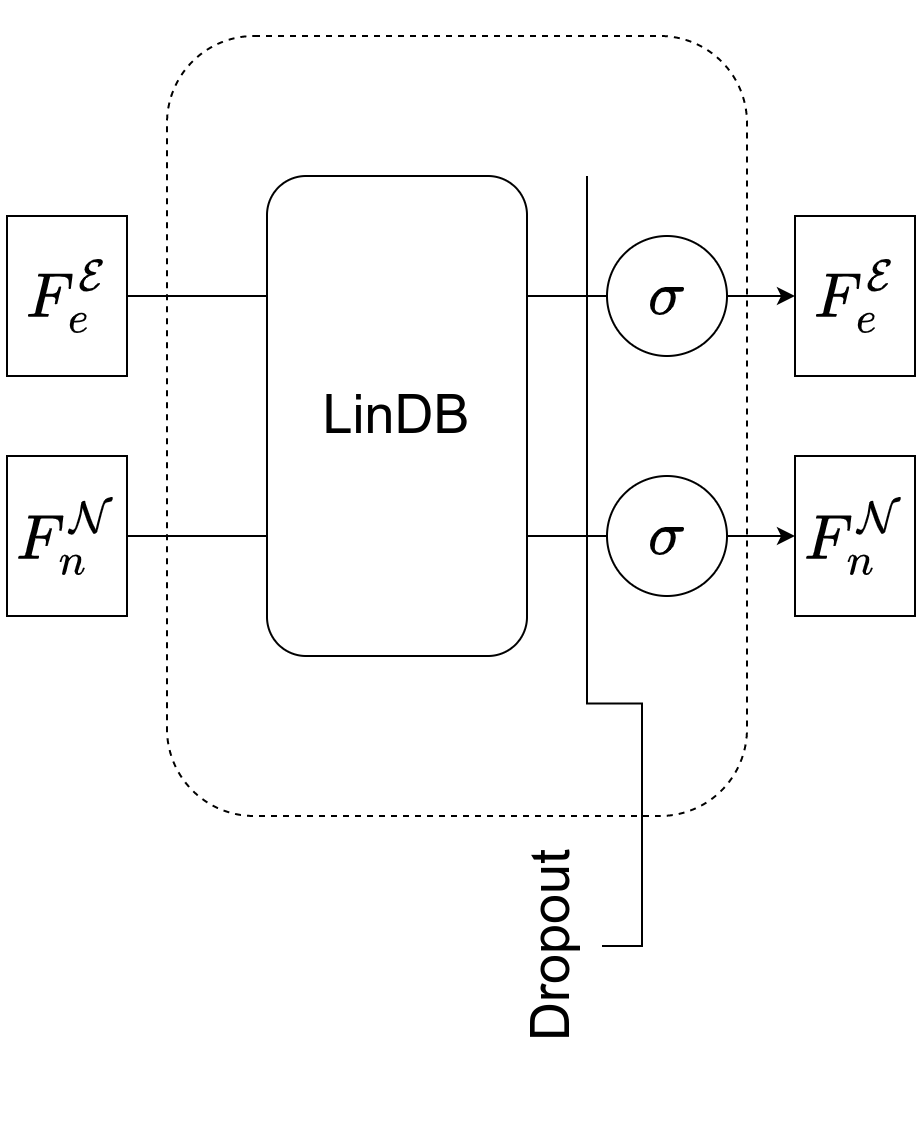}
        \caption{Dirac Bianconi 1-Step (DB1S)}
        \label{fig:DB-Operator_1step}
    \end{subfigure}
    \begin{subfigure}[b]{.25\textwidth}
        \centering
        \includegraphics[width=0.9\columnwidth]{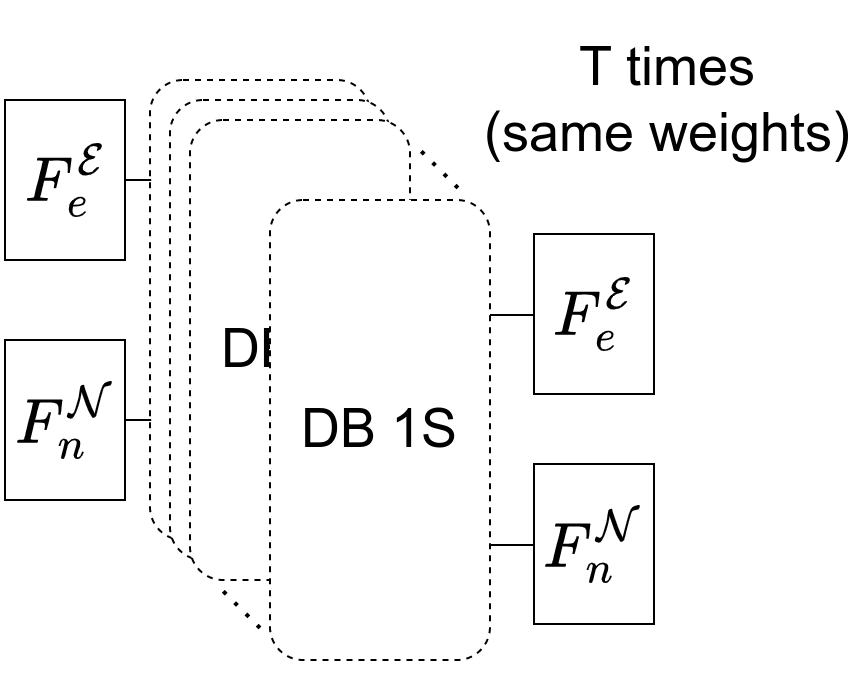}
        \caption{Dirac Bianconi T-Step}
        \label{fig:DB-Operator_Tstep}
    \end{subfigure}
    \begin{subfigure}[b]{.48\textwidth}
        \centering
        \includegraphics[width=\textwidth]{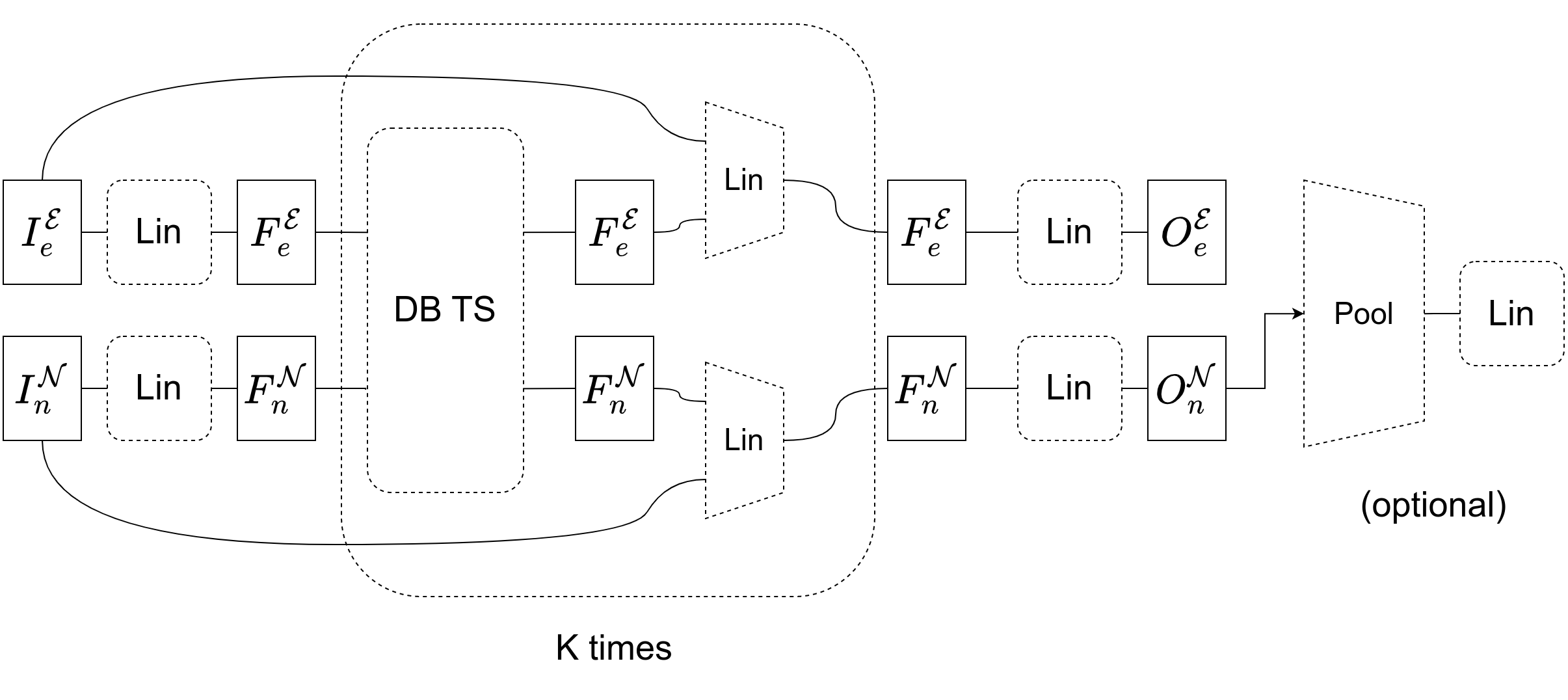}
            \caption{DBGNN layer, where Lin denotes a linear layer.}
        \label{fig:DBGNN}
    \end{subfigure}
    \vspace{-.3cm}
     \caption{Illustration of the Dirac--Bianconi Graph Neural Network (DBGNN) layer (c) and its components. The Dirac--Bianconi T-step layer (b) consists of multiple DB1S (a). By stacking multiple DB1S and applying them sequentially, information can be propagated along the graph.}
    \label{fig:DB-Operator}
\end{figure*}

We define the DB 1-Step layer as one step of the linear DB \Cref{eq:linDB}, followed by a dropout and a non-linearity. The matrices  $\mW^{ne} \in \sR^{d_n \times d_e}$, $\mW^{en} \in \sR^{d_e \times d_n}$, $\mW^{n}_\beta \in \sR^{d_n \times d_n}$, and $\mW^{e}_\beta \in \sR^{d_e \times d_e}$ are learnable weights. This layer is sketched in \Cref{fig:DB-Operator_1step}. We sequentially apply $T$ such layers with shared weights to obtain the DB T-step layer of \Cref{fig:DB-Operator_Tstep}. The full Dirac Bianconi Graph Neural Network (DBGNN) with $K$ T-step layers is then constructed as in \Cref{fig:DBGNN}:
\begin{itemize}
\vspace{-.3cm}
    \item First we linearly map the input features to the hidden feature spaces $F_n = \sR^{{d_n^\text{hidden}}}$ and $F_e = \sR^{d_e^\text{hidden}}$ for nodes and edges, respectively. \vspace{-.3cm}
    \item We then alternate DB T-step layers and skip connections that mix in the input features using a linear map $K$ times. This allows for different dynamics that see both the initial conditions and the features processed by the previous layers.\vspace{-.3cm}
    \item Finally, we use MLP to map from the hidden feature dimension to the output dimension, optionally followed by pooling and another MLP layer.\vspace{-.3cm}
\end{itemize}
Such a DBGNN makes $KT/2$ node-to-node hops on the graph. Before testing DBGNNs on real datasets, we perform synthetic analyses that demonstrate the key properties, including long-range capabilities.

\section{Synthetic analysis of DBGNN properties}
\textbf{DBGNNs do not suffer from oversmoothing}
\label{sec_syntheticAnalysis_NoOversmoothing}
A common way to understand whether an architecture suffers from oversmoothing is to study the different variations of the Dirichlet energy (DE) \citep{zhouDirichletEnergyConstrained2021,wangACMPAllenCahnMessage2023,ruschGraphCoupledOscillatorNetworks2022,chenDirichletEnergyEnhancement2023,liuFrameletMessagePassing2023,fuImplicitGraphNeural2023,digiovanniUnderstandingConvolutionGraphs2023}, for which the definition and computation are given in \Cref{app:dirichlet_energy}. 

To assess the intrinsic equilibration properties of DBGNNs in comparison to Graph Convolutional Networks (GCNs), we evaluate the DE over approximately 1,000 steps for DBGNNs and 100 steps for GCNs in untrained networks. GCNs are included in our analysis due to their widespread use and established effectiveness in various applications.

\begin{figure*}[h]
    \centering
    \begin{subfigure}[b]{.4\textwidth}
        \includegraphics[width=\textwidth]{"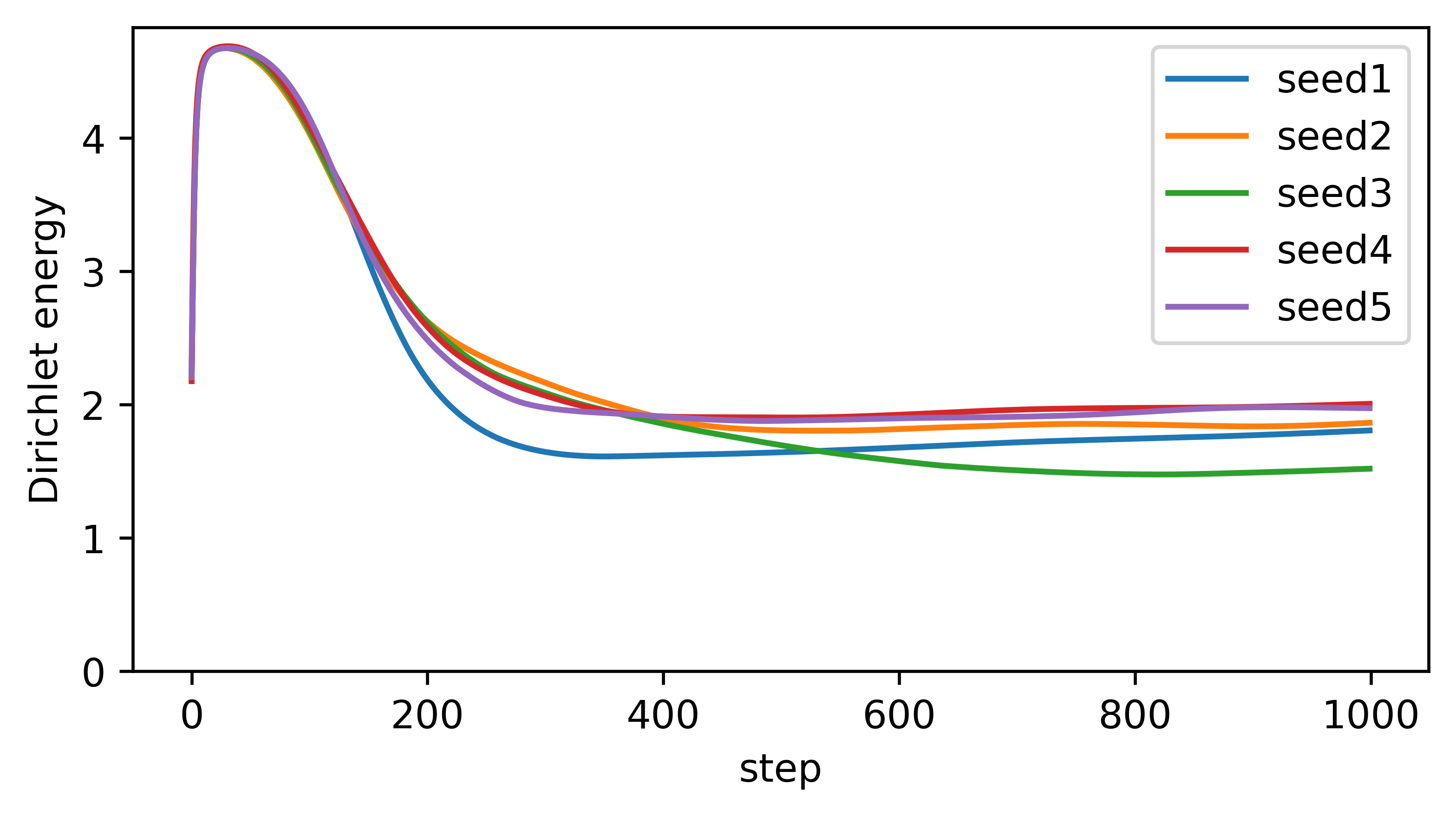"}    
        \caption{DBGNN with 1 layer and 1,000 steps per layer}
    \end{subfigure}
    \begin{subfigure}[b]{.4\textwidth}
        \includegraphics[width=\textwidth]{"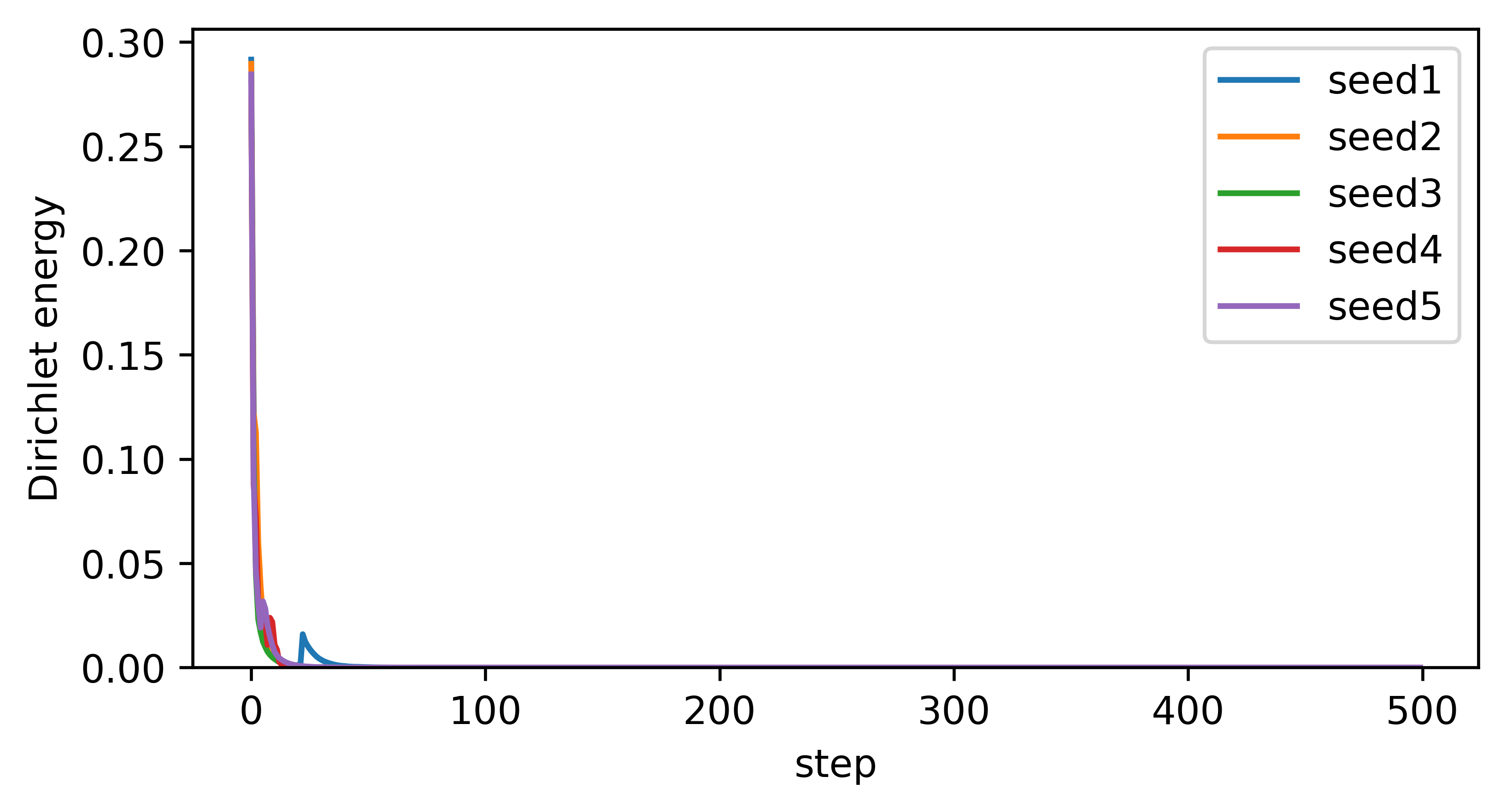"}
        \caption{GCN with 100 layers}
    \end{subfigure}
    \caption{Evolution of the normalized Dirichlet energy (DE) of the node feature embeddings for a sample of dataset20 with five different seeds and no training. For DBGNNs, the DE remains at a high level, meaning that information can be deeply propagated, while for GCNs, information is quickly lost.}
    \label{fig:dirichlet_energy_init}
\end{figure*}

The results are shown in \Cref{fig:dirichlet_energy_init}. From the spectral analysis of \Cref{eq:top-DB}, we find that no equilibration occurs for DBGNNs, while GCNs introduced by \citet{kipfSemiSupervisedClassificationGraph2017} quickly lose heterogeneity. Hence, DBGNNs allow information to be propagated across very long ranges, while in the case of GCNs, information is diffused quickly. \looseness-1

\textbf{Wave aspects of DBGNNs enable deep propagation of signals}
To analyze the long-range capabilities of DBGNNs, we analyze how the layers can spread a localized feature into a graph. Compared to MPNNs, there are two aspects that might enable deep spreading: One is the intrinsic wave dynamics of the linear DB equation, and the other is that we apply a non-linearity to the edges. This second aspect resembles the approach of \citep{bodnarNeuralSheafDiffusion2022,zhaoGraphNeuralConvectionDiffusion2023,rossiEdgeDirectionalityImproves2023}. It is noteworthy that with a ReLU activation function on the edges, either $x_i - x_j$ or $x_j - x_i$ is completely suppressed. This induces a strong directionality in the behavior of the layer, which could enable long-range propagation. 

To investigate the effect of edge non-linearity and edge updating separately, we will compare \Cref{eq:linDB} and the iterated DB 1-step layer with \Cref{eq:MPNN} and an MPNN \Cref{eq:MPNN-sigma} with and without edge non-linearity:
\begin{align}
    x_{i}(t+1) &= \sigma(\mW^{ne} \sum_{j \in \gN(i)} e_{ij} + \mW_\beta^{n} x_{i}(t))\label{eq:MPNN-sigma}\\
    e_{ij}(t) &= \sigma(\mW^{en} (x_i(t) - x_j(t)))\nonumber
    \quad \text{   or   } \quad \\
    e_{ij}(t) &= \mW^{en} (x_i(t) - x_j(t)), \nonumber
\end{align}

for the same randomly initialized weights drawn from a normal distribution with spread $0.1$. As discussed in the introduction, \Cref{eq:linDB} can induce both oscillatory and non-oscillatory behavior, depending on the weights. Here we investigate the properties of the oscillatory regime, in which we might expect propagating waves. To do so, we constrain the weights to be $\mW^{ne} = - {\mW^{en}}^\dagger$ and $\mW^{n/e}_\beta$ antisymmetric. \looseness-1
\begin{figure*}
    \centering
    \includegraphics[width=.3\textwidth]{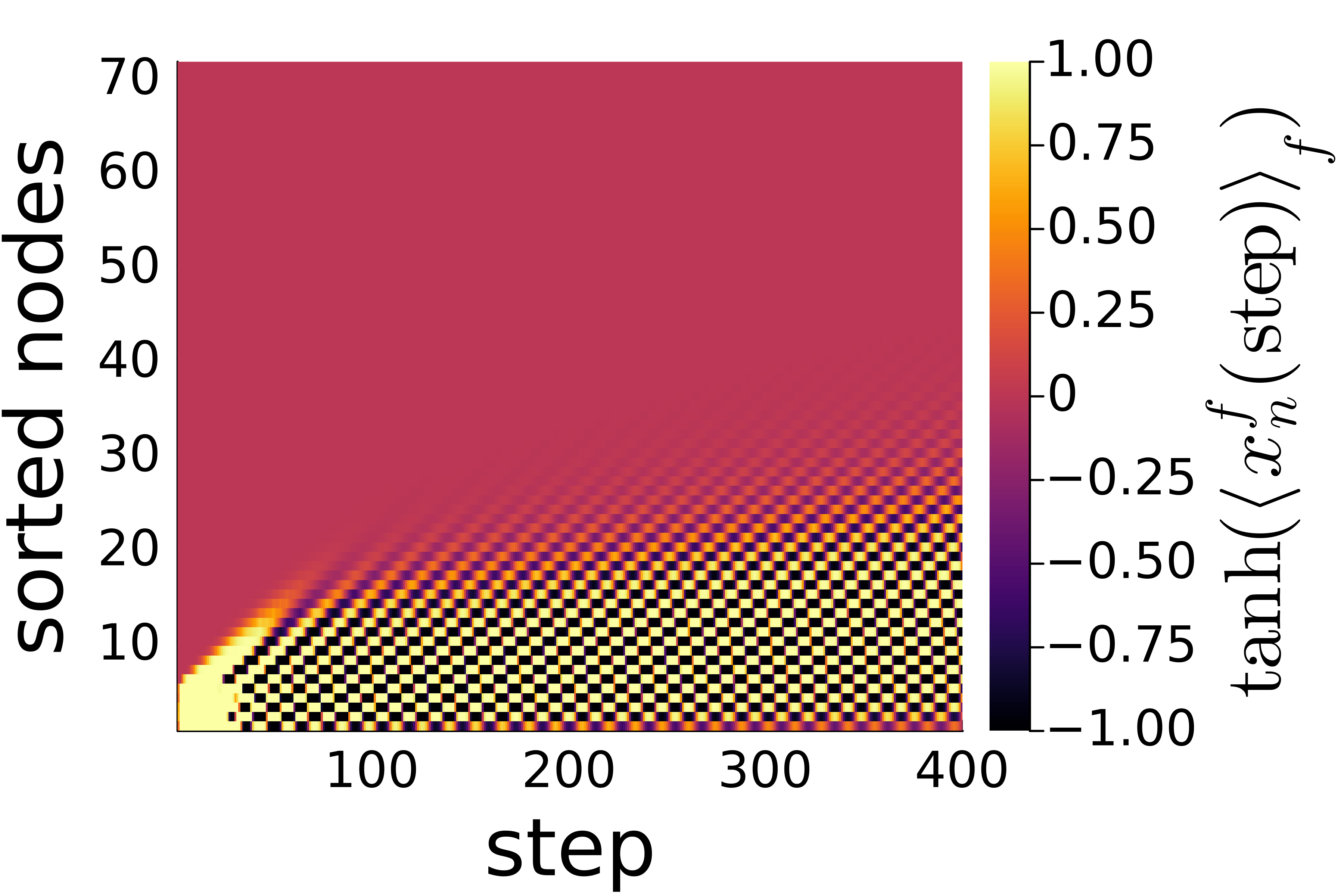}
    \includegraphics[width=.3\textwidth]{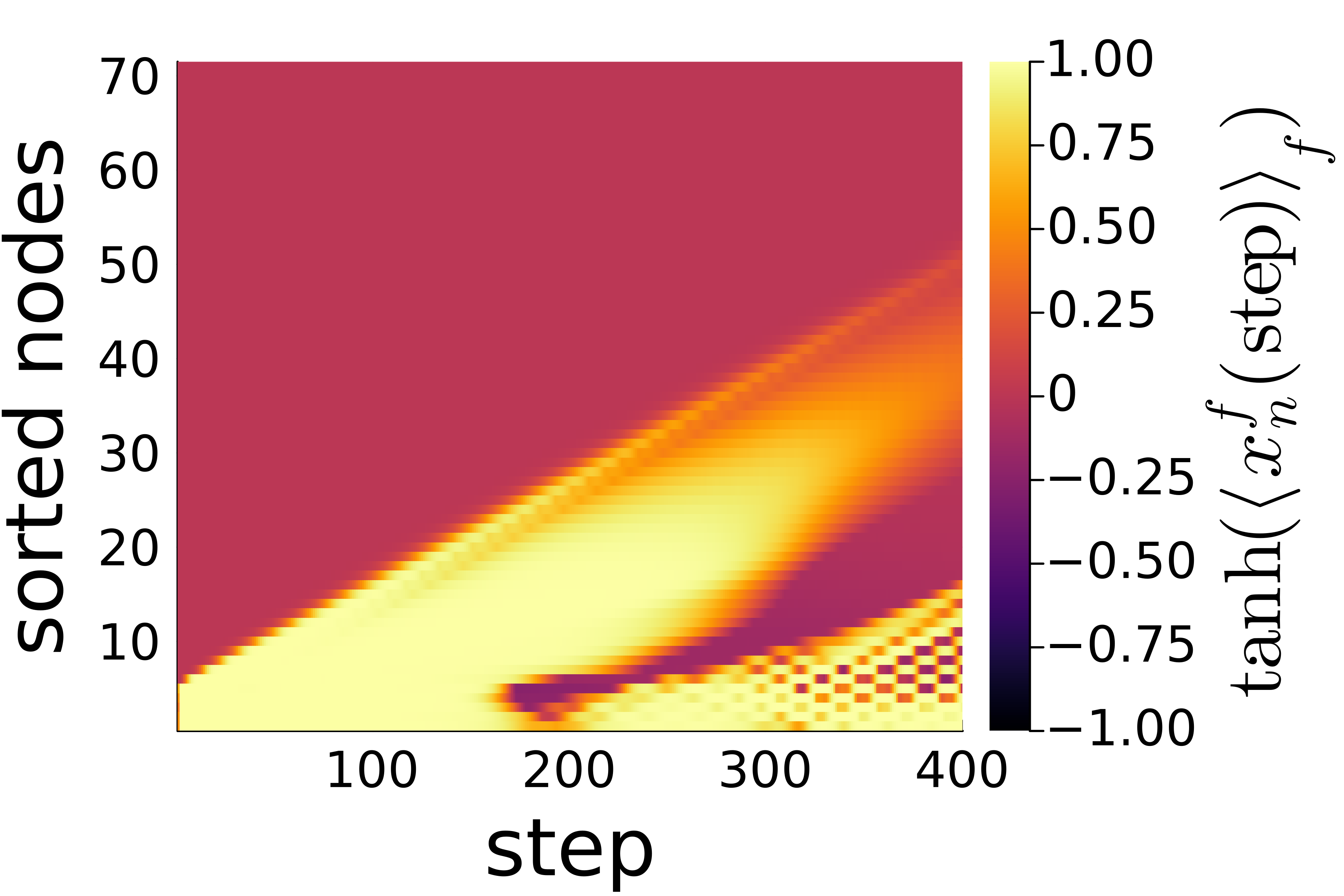}
    
    \includegraphics[width=0.3\textwidth]{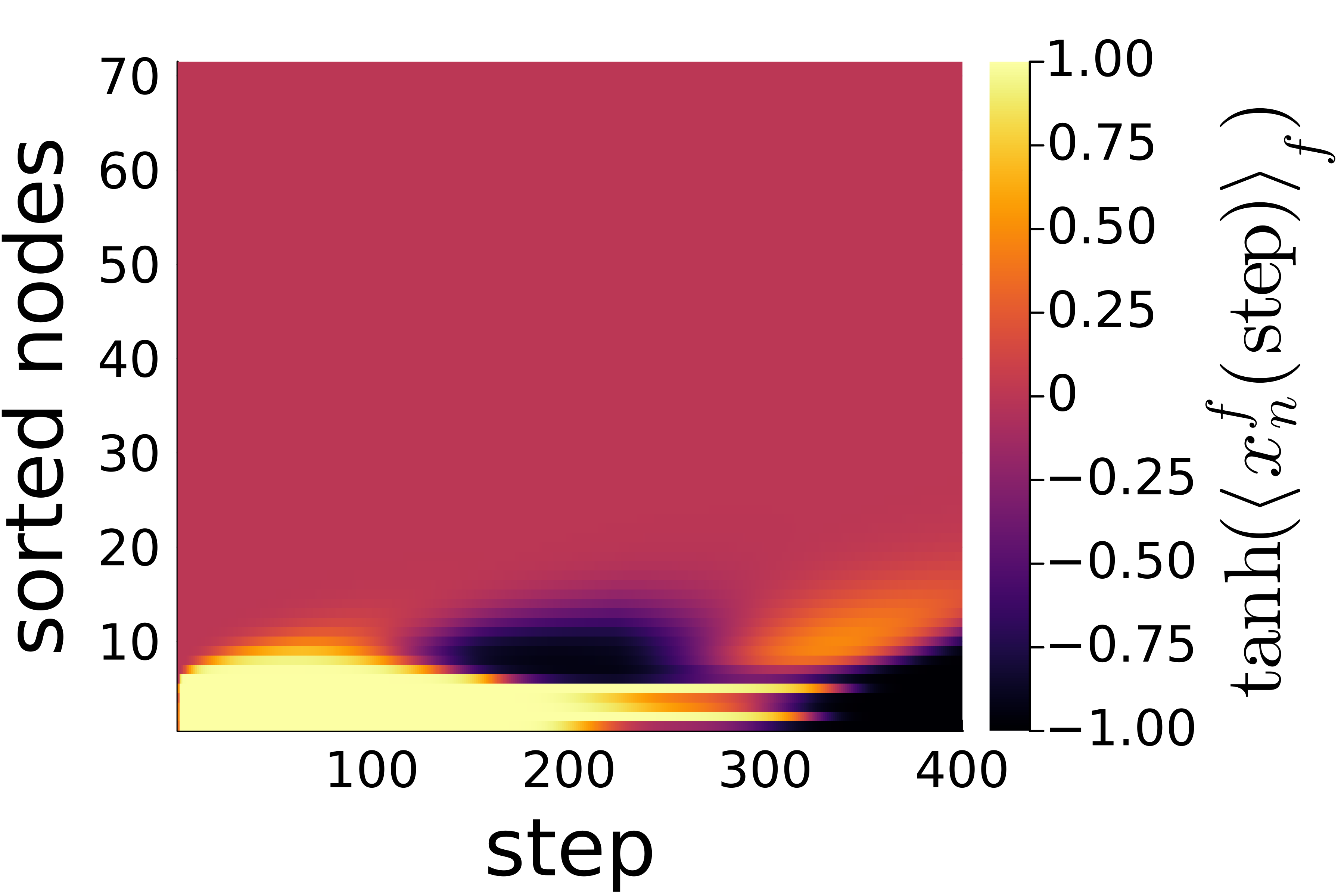}
    \includegraphics[width=0.3\textwidth]{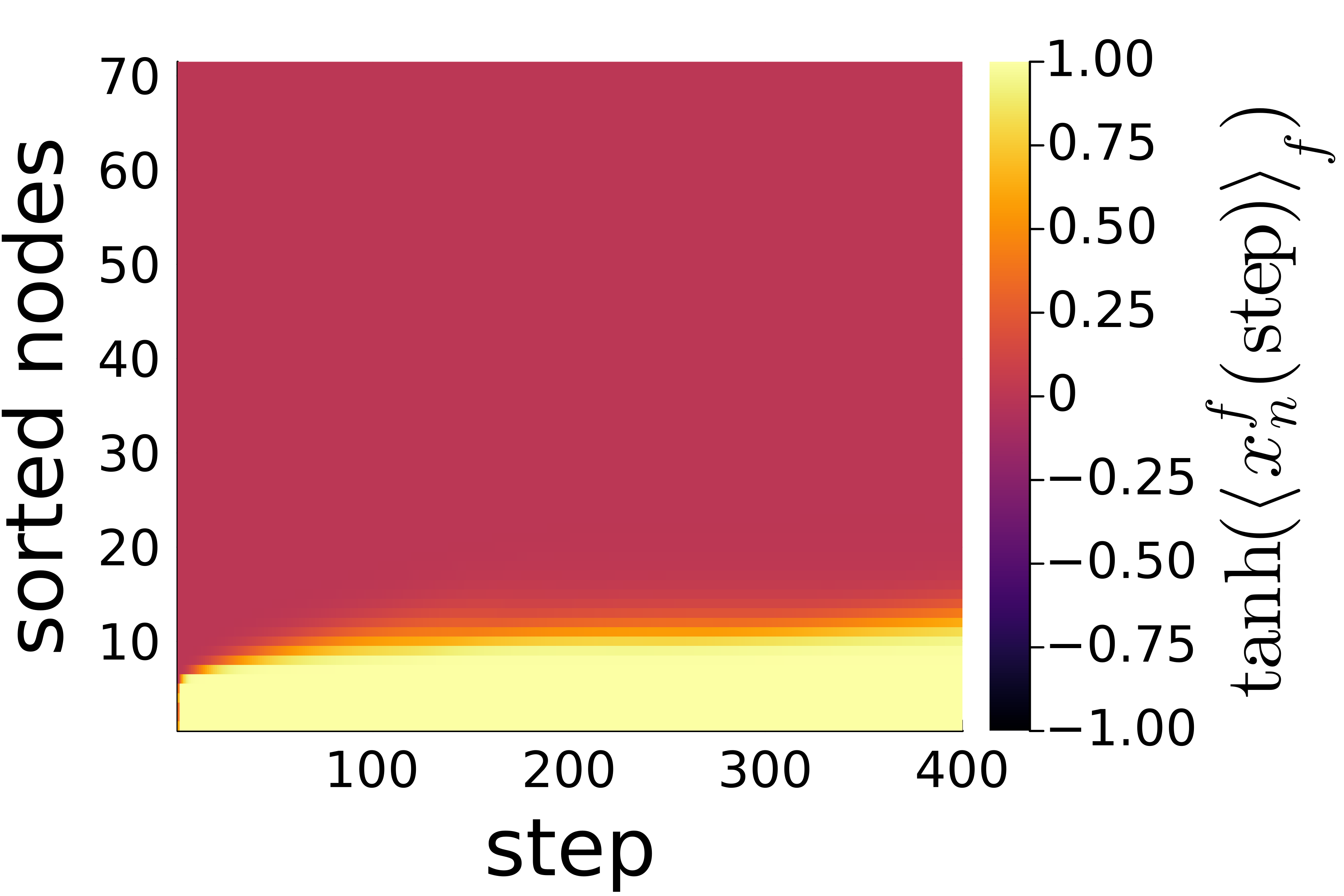}
    \includegraphics[width=0.3\textwidth]{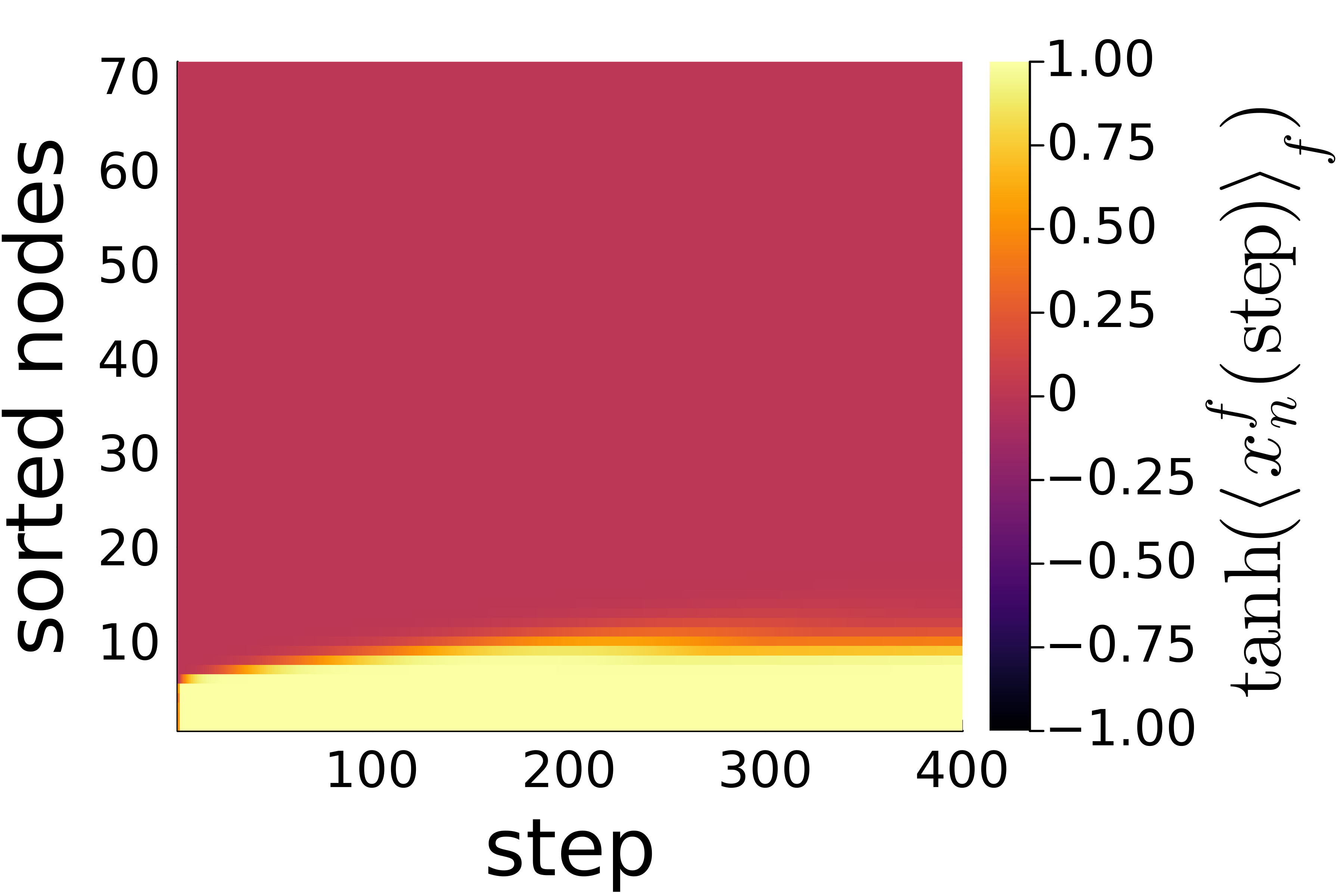}
    \caption{Oscillatory regime: Feature activation versus steps of the linear DB \Cref{eq:linDB} (top left), the non-linear DB 1-step layer \Cref{eq:linDB} + ReLU (top right), the linear MPNN layer \Cref{eq:MPNN} (bottom left), and MPNN \Cref{eq:MPNN-sigma} without (middle) and with non-linear messages (bottom right). $d_n = d_f = 4$, same random weights. In the case of DBGNN, information can be propagated and is not lost by diffusion.}
    \label{fig:spreading-oscil}
    \vspace{-.3cm}
\end{figure*}

We evolve these models on a 5x20 rectangular grid, where one of the short edges has all nodes randomly initialized, and all other edge and node features are identically zero. \Cref{fig:spreading-oscil} shows example trajectories for five models with four edge and node features to illustrate the behavior of DBGNNs in comparison to MPNNs. The linear DB equation shows a leading edge, a concentrated wave of activation that spreads rapidly into the network before dissipating, with ripples radiating into the rest of the network. Due to the oscillatory initialization, the linear MPNN equation also shows oscillatory behavior, but this does not result in spreading into the network. Adding the non-linearities stabilizes the leading edge of the DB equation, which now reaches the other end of the graph, and also sharpens the ripples into a coherent excitation that travels slower down the graph. For the MPNN, the non-linearities suppress the oscillations, leaving us with pure diffusion. For higher dimensional internal spaces, as well as for many non-oscillatory random weights, most configurations of all layers exhibit slow diffusion in the system. Occasionally, we can randomly generate coherent traveling excitations in the DBGNN, which are not observed in the MPNN. We provide examples of these trajectories in the appendix. We conclude that the wave aspects of the DBGNN enable deep propagation of signals into the graph, while edge non-linearities play a minor role. \looseness-1 %

\section{Experimental results}
We evaluate the performance of our DBGNN on challenging tasks involving long-range dependencies. The datasets deal with power grid properties and properties associated with the molecular structures of peptides. Power grids are known to encompass long-range dependencies, as highlighted by \citet{ringsquandlPowerRelationalInductive2021}. For the power grid dataset, GNNs capable of effectively propagating information over extended distances demonstrate superior performance \cite{nauckDynamicStabilityAssessment2023}. The emphasis for this dataset is on topological relationships, given the absence of edge features and the presence of only one binary node feature. In contrast, the molecular dataset exhibits a diverse array of node and edge features. These peptide datasets were released with the specific purpose of benchmarking the handling of long-range dependencies by various GNNs. The selected tasks encompass node regression, graph regression, and graph classification. \looseness-1

\textbf{Dynamic stability of power grids}
The most sophisticated dataset dealing with the dynamic stability of power grids is published by \citet{nauckDynamicStabilityAssessment2023}, which is based on \citet{nauckPredictingBasinStability2022,nauckDynamicStabilityAnalysis2022}. There are a total of 20,000 grids: 10,000 small grids of size 20 (dataset20), and 10,000 medium-sized grids of size 100 (dataset100). Besides training and evaluating the models on the same grid sizes, we also analyze the out-of-distribution generalization by training the models on grids of size 20 and evaluating them on grids of size 100. We follow the nomenclature from \citet{nauckDynamicStabilityAssessment2023} and refer to this task as tr20ev100. \looseness-1

Training models on smaller grids and evaluating them on larger grids is important for real-world applications because the computational effort increases at least quadratically with the size of the grid. 

The results are given in \Cref{tab:results_snbs}, where we compare them with the current benchmark performances. The DBGNN achieves the best performance on all tasks and significantly outperforms the other models on out-of-distribution generalization. One of the reasons for its superior performance may be related to its ability to go deep without encountering the problem of oversmoothing. The final DBGNN consists of 4 DB 12-step layers, resulting in a total of 48 steps. \looseness-1 

\begin{figure}[htb]
    \centering
        \includegraphics[width=\linewidth]{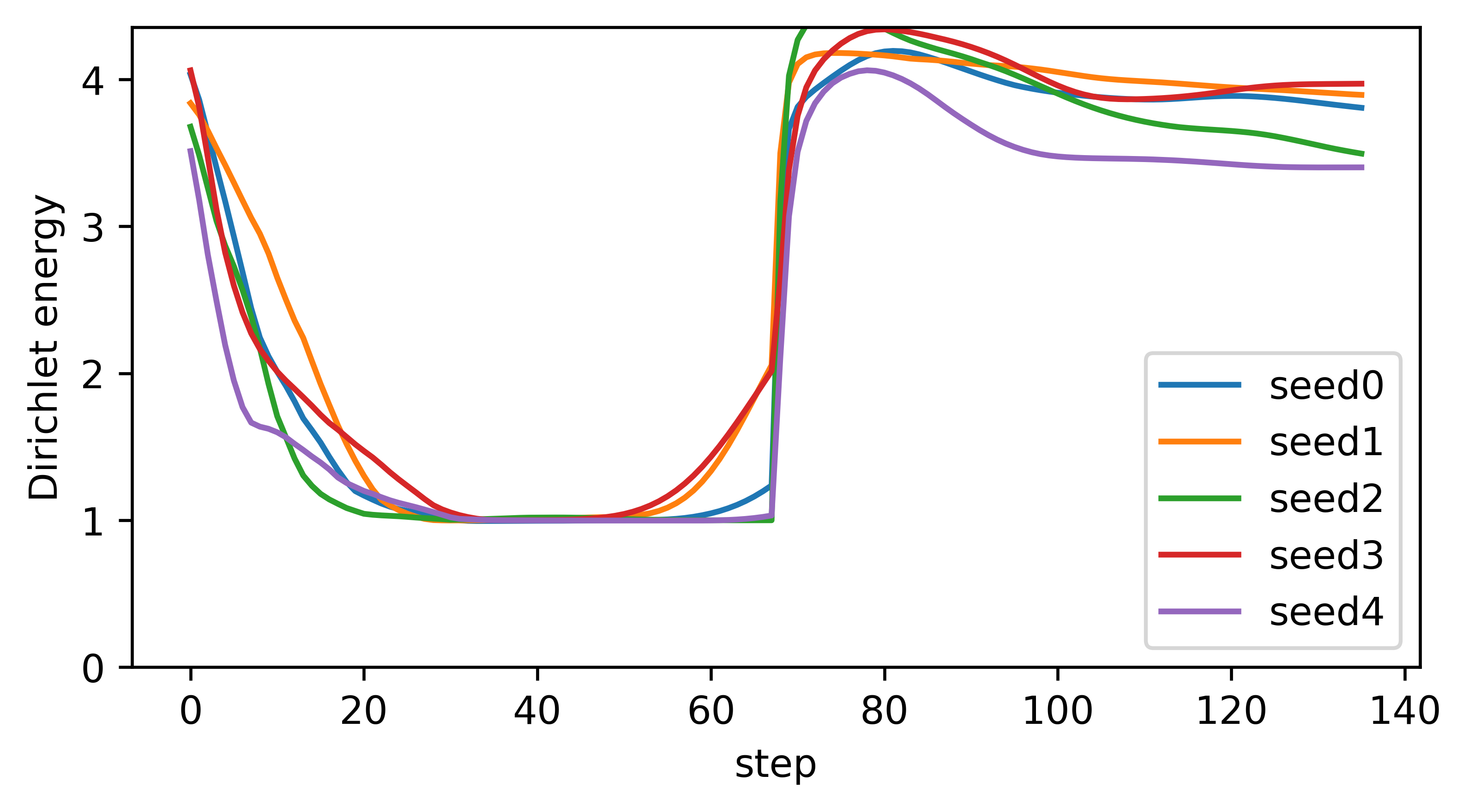}
       
    \caption{Evolution of normalized Dirichlet energy of node feature embeddings in a trained DBGNN layer for a sample of dataset20 with five different seeds.}
    \label{fig:dirichlet_energy}
    \vspace{-.3cm}
\end{figure}

To further investigate the absence of smoothing, we compute the Dirichlet energy for a sample of dataset20 in the forward pass using the node embeddings at each step. \Cref{fig:dirichlet_energy} shows that for trained models, the Dirichlet energy remains high throughout the forward pass, confirming the intuition that DBGNNs do not suffer from oversmoothing even at considerable depth. Some seeds go through periods of considerable "sharpening", especially following the change in  dynamics after each T-step. \looseness-1 

\begin{table}
    \small
	\centering
	\caption{Performance of dynamic stability prediction measured by the $R^2$ score in \% using the benchmark models from \citet{nauckDynamicStabilityAssessment2023}, distinguishing between in- and out-of-distribution (distr.) The columns \textit{tr20ev20} and \textit{tr100ev100} indicate that the models are trained and evaluated on the same datasets. The out-of-distribution performance is measured by evaluating the models on dataset100 after training them on dataset20 (tr20ev100). Following \citet{nauckDynamicStabilityAssessment2023}, we show the mean performance of the best three configurations out of five different initializations.}
 \begin{tabularx}{\linewidth}{XYYYYY}
		\toprule
     \textbf{Model}   & \multicolumn{2}{c}{\textbf{In-distr.}} & \multicolumn{1}{c}{\textbf{Out-of-distr.}} \\
 \cmidrule(r){1-1}  \cmidrule(rl){2-3} \cmidrule(l){4-4}
	     & tr20ev20 & tr100ev100 & tr20ev100\\
 \cmidrule(r){1-1}  \cmidrule(rl){2-3} \cmidrule(l){4-4}
        ArmaNet &  82.22 \tiny{$\pm$ 0.12}&  88.35 \tiny{$\pm$ 0.12} & 67.12 \tiny{$\pm$ 0.80} \\
        GCNNet  & 70.74 \tiny{$\pm$ 0.15} & 75.19 \tiny{$\pm$ 0.14} & 58.24  \tiny{$\pm$ 0.47} \\
        TAGNet & 82.50 \tiny{$\pm$ 0.36} & 88.32 \tiny{$\pm$ 0.10} & 66.32  \tiny{$\pm$ 0.74} \\ 
 \cmidrule(r){1-1}  \cmidrule(rl){2-3} \cmidrule(l){4-4}
        DBGNN & \textbf{85.68} \tiny{$\pm$ 0.10} & \textbf{90.08} \tiny{$\pm$ 0.02} & \textbf{73.73} \tiny{$\pm$ 0.07}\\
        
	 \bottomrule
 	\end{tabularx}
 	\label{tab:results_snbs}
\end{table}

\textbf{Peptide property prediction}
To assess the performance of the DBGNN model on long-range interactions, we use the \textit{Peptides-struct} dataset from the long-range benchmark dataset~\cite{dwivediLongRangeGraph2023}. Peptides, defined as short chains of amino acids, play a crucial role in numerous biological processes. Given the intricate relationships between peptides and their biological functions, computational prediction of peptide properties is crucial for advancing drug development and biomolecular engineering. The peptide datasets contain 15,535 graphs, with an average of 150.94 nodes per graph and an average diameter of 56.99.

The \textit{Peptides-struct} dataset is a multi-label graph regression task. Various properties of the peptides are predicted, with the full details presented in \citet{dwivediLongRangeGraph2023}. \looseness-1

The performance of the baselines from \citet{dwivediLongRangeGraph2023} and the DBGNN is shown in \Cref{tab:results_peptides_one_column}. The baselines make use of additional methods such as random-walk structural encoding (RWSE) \cite{dwivediGraphNeuralNetworks2022} and Laplacian positional encoding (LapPE) \cite{dwivediBenchmarkingGraphNeural2020} to improve performance. Spectral attention networks (SANs) are improved transformers introduced by \citet{kreuzerRethinkingGraphTransformers2021}. \citet{tonshoffWhereDidGap2023} show that the performance of MPNNs such as GCNs can be significantly improved by adding multi-layer regression heads to the output of GNNs. However, we refrain from including multi-head regressions in order to focus on the pure properties of message passing components. \looseness-1
The DBGNN, consisting of only about a quarter of the parameters, still significantly outperforms all message passing graph neural networks, but achieves lower performance than the transformers. We conclude that the DBGNN is a powerful MPNN capable of dealing with node and edge features and correctly identifying long-range dependencies. \looseness-1
\begin{table}
\vspace{-.2cm}
    \small
	\centering
	
    \caption{Performance on \textit{Peptides-struct} visualized by mean absolute error (MAE). The performance of the baseline is taken from \citet{dwivediLongRangeGraph2023}. The number of parameters is shown in column \# param.}
 \begin{tabularx}{\linewidth}{XYYYYY}
		\toprule
    \textbf{Model}   & \textbf{\# param} & train MAE & test MAE \\
     \midrule
        GCN &  508k & 0.2939 \tiny{±0.0055} & 0.3496 \tiny{±0.0013} \\
        GCNII & 505k & 0.2957 \tiny{±0.0025}  &  0.3471 \tiny{±0.0010}\\      
        GINE & 476k & 0.3116 \tiny{±0.0047}  & 0.3547 \tiny{±0.0045} \\
        GatedGCN & 509k &0.2761\tiny{±0.0032}&0.3420 \tiny{±0.0013}\\
        GatedGCN+RWSE &506k &0.2578 \tiny{±0.0116}  &0.3357 \tiny{±0.0006}\\
        \midrule
        Transform+LapPE & 488k &0.2403 \tiny{±0.0066}  & \textbf{0.2529} \tiny{±0.0016}\\
        SAN+LapPE & 493k & 0.2822 \tiny{±0.0108}& 0.2683 \tiny{±0.0043}\\
        SAN+RWSE & 500k & 0.2680 \tiny{±0.0038}  &  0.2545 \tiny{±0.0012}\\
        \midrule
        DBGNN & \textbf{127k} & \textbf{0.2433} \tiny{±0.0044} & 0.2864 \tiny{±0.0046}\\
       
	 \bottomrule
 	\end{tabularx}
 	\label{tab:results_peptides_one_column}
  \vspace{-.2cm}
\end{table}

\textbf{Comparison with other oversmoothing techniques}
Considering the inherent flaw of oversmoothing in GNNs and especially in MPNNs, various methods have been introduced to mitigate the issue \cite{ruschSurveyOversmoothingGraph2023}. These approaches include diverse strategies such as applying normalization techniques, employing regularization methods, incorporating residual connections, and modifying the dynamics of GNNs \cite{ruschSurveyOversmoothingGraph2023}. Residual connections, exemplified by models such as GatedGCN \cite{bressonResidualGatedGraph2018} and GCNII \cite{chenSimpleDeepGraph2020}, provide a way to address oversmoothing. GCNII is specifically designed to tackle oversmoothing using \textbf{I}nitial residual and \textbf{I}dentity mapping. Distinct but related methods include ArmaNet \cite{bianchiGraphNeuralNetworks2021} and TAGNet \cite{duTopologyAdaptiveGraph2017}, which involve multiple steps within a single layer, enabling the construction of deeper GNNs that provide meaningful output. To address oversmoothing in  MPNNs, \citet{kreuzerRethinkingGraphTransformers2021} propose SANs and use positional encoding techniques.\looseness-1

In contrast, the DBGNN motivated by dynamical systems has several interesting implications. One is that it is inherently non-oversmoothing, as demonstrated in \Cref{sec_syntheticAnalysis_NoOversmoothing}. This intrinsic attribute potentially contributes to the outstanding performance of the DBGNN across various datasets, as demonstrated in \Cref{tab:results_snbs,tab:results_peptides_one_column}, which may provide valuable insights for
the development of enhanced GNN models.\looseness-1
\section{Conclusion}
We introduced a new graph neural network layer based on a straightforward generalization of the topological Dirac--Bianconi equation on networks. We show that this model has no intrinsic tendency to equilibrate features on the network. This has the potential to handle long-range dependencies and treat edge and node features equally. By incorporating multiple steps with weight sharing within a layer, we enable the layer to efficiently learn dynamics that probe the graph deeply. The DBGNN is a straightforward adaptation of the topological Dirac--Bianconi equation, and thus offers the potential for many further modifications. In its current form, the DBGNN already outperforms other layers in predicting the dynamic stability of power grids and achieves competitive performance in predicting molecular properties, outperforming conventional MPNN methods.\looseness-1

By analyzing the internal node embeddings using the Dirichlet energy, we can show that the DBGNN does not seem to suffer from the oversmoothing problem. Furthermore, we observe a sudden sharpening of the features when the dynamics change after many steps, a phenomenon that needs to be better understood. We provide evidence that the long-range capabilities result from the underlying Dirac--Bianconi dynamics rather than from edge non-linearity. 
The experiments were performed without adapting the model to the task at hand. Since the DBGNN is close to standard MPNN-style networks, the vast array of modifications that exist for them can be easily adapted here. It remains to be seen whether they can also enhance DBGNN-style networks. One particularly interesting question is whether long-range modifications to GCNs, such as \citet{gutteridgeDRewDynamicallyRewired2023}, can further enhance the long-range behavior of DBGNNs. \looseness-1

Overall, the expanded long-range capabilities provide new opportunities for power grid analysis and molecular prediction, offering increased potential for scientific exploration and understanding. \looseness-1

\section{Acknowledgements}
All authors gratefully acknowledge Land Brandenburg for supporting this project by providing resources on the high-performance computer system at the Potsdam Institute for Climate Impact Research. The work was in part supported by DFG Grant Numbers KU 837/39-2 (360460668), BMWK Grant 03EI1016A, and BMBF Grant 03SF0766. Christian Nauck would like to thank the German Federal Environmental Foundation (DBU) for funding his PhD scholarship and Professor Jörg Raisch for supervising his PhD. Michael Lindner greatly acknowledges support from the Berlin International Graduate School in Model and Simulation (BIMoS) and from his doctoral supervisor Professor Eckehard Schöll. Rohan Gorantla was supported by the United Kingdom Research and Innovation (grant EP/S02431X/1), UKRI Centre for Doctoral Training in Biomedical AI at the University of Edinburgh, School of Informatics and Exscientia Plc, Oxford. AI tools are used on the (sub-)sentence level to improve language. We especially want to thank the reviewers for their valuable comments and feedback, which have significantly enhanced the quality of this paper. We also express our gratitude to Teresa Gehrs from LinguaConnect and the Proofreading Service for Junior Scholars at TU Berlin for their help in proofreading this paper.

\FloatBarrier


\newpage
\appendix
\onecolumn
\section{Appendix}
\subsection{Data availability}
The code to train the DBGNNs and to generate the figures is available on \url{https://github.com/PIK-ICoNe/DBGNN_paper-companion.git} and \url{https://doi.org/10.5281/zenodo.12687981}.

\subsection{Dirichlet energy}
\label{app:dirichlet_energy}
The Dirichlet energy is a measure of the heterogeneity of features across the graph. The normalized Dirichlet energy (DE) is computed by:
\begin{equation}
    \text{Dirichlet energy} = \frac{\text{tr} ({x{(k)}}^\top \mL x{(k)})}{\text{tr}({x{(k)}}^\top  x{(k)})}, 
\end{equation}
where $x{(k)}$ denotes the node embedding after $k$ steps and $\text{tr}$ denotes the trace operator.
The interpretation becomes apparent by rewriting the Dirichlet energy in terms of edge differences:\looseness-1
\begin{equation}
    \text{Dirichlet energy} = \frac{\sum_{(i,j) \in \gE} ||x^i(k) - x^j(k)||^2}{\sum_{i \in \gN} ||x^i(k)||^2}, 
\end{equation}
where $x^i(k) \in F_n$ denotes the state of node $i$ after $k$ steps.

\subsection{Hyperparameters for reproducibility}
The following tables provide the hyperparameters to reproduce the main results. For the power grid dataset, the information is given in \Cref{tb:snbs_parameters}, and for the peptide structure task in \Cref{{tb:peptide_structure_parameters}}.

\begin{table}[htb]
	\centering
	\caption{Properties of DBGNN after hyperparameter studies for power grid datasets}
    \begin{tabularx}{\linewidth}{XXX}
    \toprule
        parameter & dataset with grids of size 20 & dataset with grids of size 100 \\
		\midrule
        number of layers (K) & 2 & 2\\
        number of steps (with shared weights) per layer (T) & 68 & 68\\
        $d_n^{\text{hidden}}$ & 113 & 113 \\
        $d_e^{\text{hidden}}$ & 109 & 109 \\
        dropout for node convolution & $\approx 1.4 \times 10^{-2}$ & $\approx 2.1 \times 10^{-2}$ \\
        dropout for edge convolution & $\approx 1.9 \times 10^{-3}$ & $5.7 \times 10^{-4}$\\
        batch size & 50 & 50 \\
        epochs & 2 000 & 2 000\\
        learning rate (LR) & max $6.1 \times 10^{-4}$ & max: $\approx 2.6 \times 10^{-3}$\\ 
        scheduler & oneCycleLR with initial div factor: 32, final div factor: $ 5.8 \times 10^5$ & oneCycleLR with initial div factor: 75, final div factor: $\approx 1.7 \times 10^6$\\
	 \bottomrule
 	\end{tabularx}
	\label{tb:snbs_parameters}
\end{table}

\begin{table}[htb]
	\centering
	\caption{Properties of the DBGNN after hyperparameter studies for peptide structure datasets}
    \begin{tabularx}{\linewidth}{XX}
    \toprule
        parameter &  \textit{peptides-struct}\\
		\midrule
        number of layers (K) & 2\\
        number of steps (with shared weights) per layer (T) & 68\\
        $d_n^{\text{hidden}}$ & 113\\
        $d_e^{\text{hidden}}$ & 109\\
        dropout for node convolution &  $\approx 0.0554$ \\
        dropout for edge convolution &  $\approx 0.01058$\\
        batch size & 60\\
        epochs &  2000\\
        max. learning rate (LR) & $\approx  0.00548$\\ 
        oneCyle scheduler & init div factor: 151, final div factor: $\approx 4.217 \times 10^9 $\\
	   \bottomrule

 	\end{tabularx}
	\label{tb:peptide_structure_parameters}
\end{table}

\subsection{Implementation and computation details}
The packages \emph{GraphNeuralNetworks.jl}\citep{lucibelloGraphNeuralNetworksJl2023} and \emph{Flux.jl} \citep{innesFashionableModellingFlux2018} are used for the Julia implementation of the DBGNN \citep{bezansonJuliaFreshApproach2017}. Furthermore, \emph{Cuda.jl} \citep{besardEffectiveExtensibleProgramming2019} and \emph{MLDatasets.jl} are used. We also provide a PyTorch \cite{paszkePyTorchImperativeStyle2019} implementation using PyTorch Geometric \cite{feyFastGraphRepresentation2019}. We run all experiments on NVIDIA V100 and H100 accelerators.

For the power grid task and tr20ev20, training five seeds in parallel on one V100 takes about three days and 20 hours for 2,000 epochs, but significant overfitting occurs after roughly 500 epochs. For tr100ev100, overfitting occurs after roughly 600 epochs, and training two seeds in parallel take four days and eight hours for 2,000 epochs. For peptides-struct, 2,000 epochs with two seeds in parallel take roughly five days and 13 hours. For peptides-func, training two seeds in parallel takes about one day and 18 hours in total for 500 epochs.

\subsection{Computational complexity of the DBGNN}
The computational complexity is mostly determined by the repeated application of LinDB from \Cref{eq:linDB}, which we will focus on in the following. Hence, we neglect the mapping of the input features to the hidden feature spaces, which only needs to be done once before applying the DBGNN steps, as well as the concatenation of matrices and the application of non-linearities. For the expensive LinDB layer, we make use of sparse operators, such as PyTorch Scatter \cite{paszkePyTorchImperativeStyle2019}, by exploiting the sparsity of the adjacency matrix to reduce the computational effort, 
which is explained in more detail in \cite{blakelyTimeSpaceComplexity2021}. The following paragraph describes the computational complexity in more detail.

For simplicity, we assume that $d_e=d_n$, meaning that the hidden dimension of node and edge features is the same. There are two relevant terms of the sum for node and edge propagation, respectively. First, for node propagation, using the coupling matrix $\mW^{ne}$ and multiplying it by the edge features $F_e^\mathcal{E}$ for all neighbors yields: $\mathcal{O}(|E| \times d_n)$. The second term, which is a dense matrix multiplication of $W_\beta^n$ and the node features $F_n^\mathcal{N}$ yields: $\mathcal{O}(d_n \times d_n \times |E|)$. Similarly, for the propagation of edges, we have $\mathcal{O}(|N| \times d_n)$ and the multiplication of $\mW^{e}_\beta \times F_e^\mathcal{E}$ yields: $\mathcal{O}(d_e \times d_e \times |E|)$. These multiplications are repeated $T$-times, where $T$ denotes the number of DB steps. Hence, we get: $\mathcal{O}\left( T(|E| \times d_n + |N| \times d_n  + d_n^2 \times |E| +  d_n^2 \times |N|) \right)$, which is comparable to the effort of the GCN \cite{blakelyTimeSpaceComplexity2021}.

The runtime of the implemented layer can be improved by utilizing the sparsity of the adjacency matrix for the matrix multiplication as well, which does not work in the current implementation.

\subsection{Experimental details on power grids}
The dynamics of power grids is characterized by complex collective phenomena that extend over the whole system \citep{witthautCollectiveNonlinearDynamics2022}. The chosen task is based on the so-called single-node basin stability, originally introduced by \cite{menckHowBasinStability2013}, which describes nodal dynamic stability. It is the result of expensive Monte Carlo simulations and quantifies the probabilistic behavior of the entire power grid after applying nodal perturbations.

For the power grid models, the nodal input features are categorical representations of sources or sinks. The power lines are assumed to be homogeneous; therefore, the input edge features are simply set to 1. The absence of diverse features puts the emphasis on topological properties.

The datasets contain individual training, validation, and test sets (70:15:15). The only input feature per node, which describes whether a node is considered a source ($P=1$) or sink ($P=-1$), is based on the injected power $P$. Since homogeneous coupling is used, there are no edge features. Performance on the nodal regression setup is evaluated using the coefficient of determination ($R^2$).\looseness-1

\subsection{Non-oscillatory random weights}
\begin{figure}
    \centering
    \includegraphics[width=.3\columnwidth]{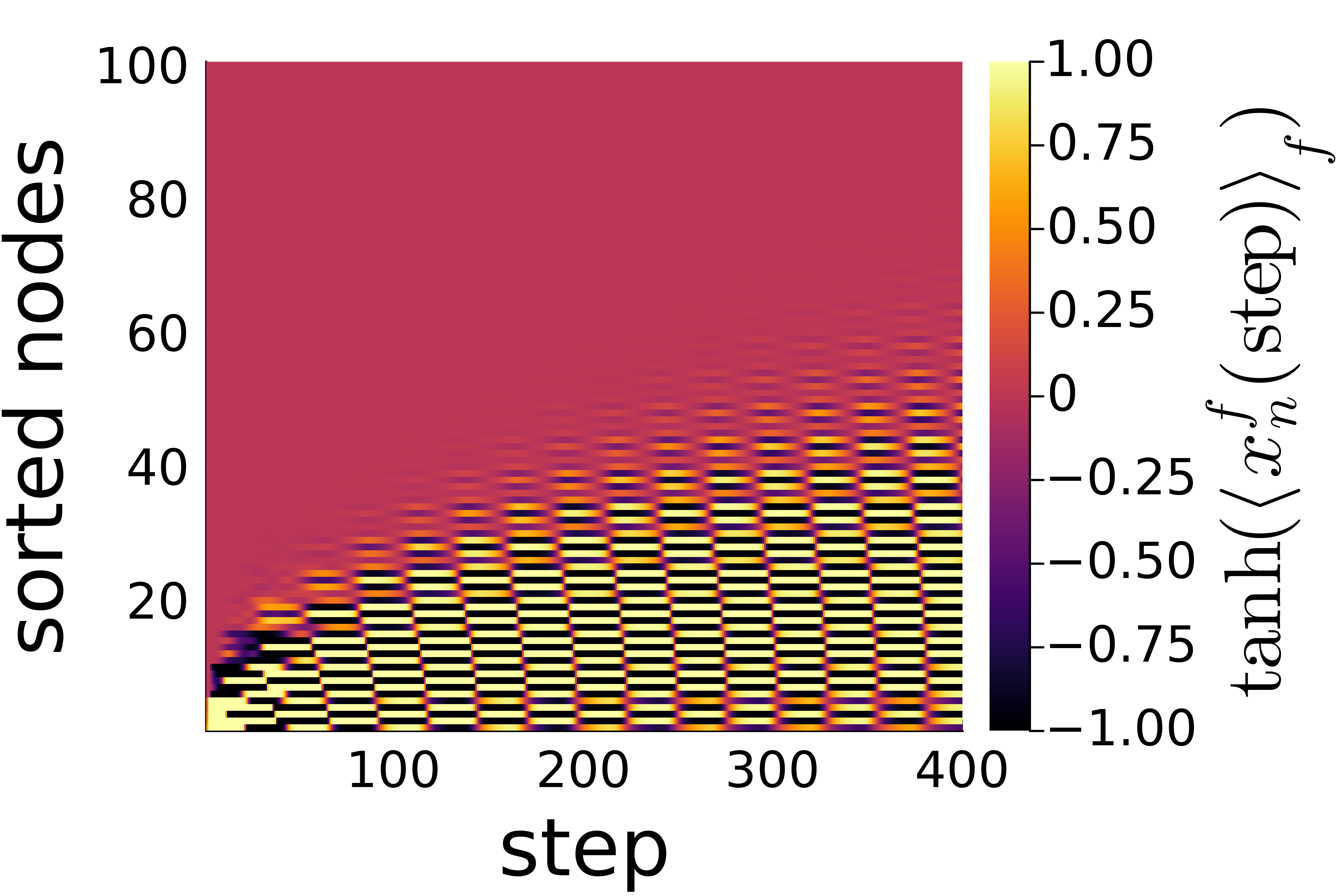}
    \includegraphics[width=.3\columnwidth]{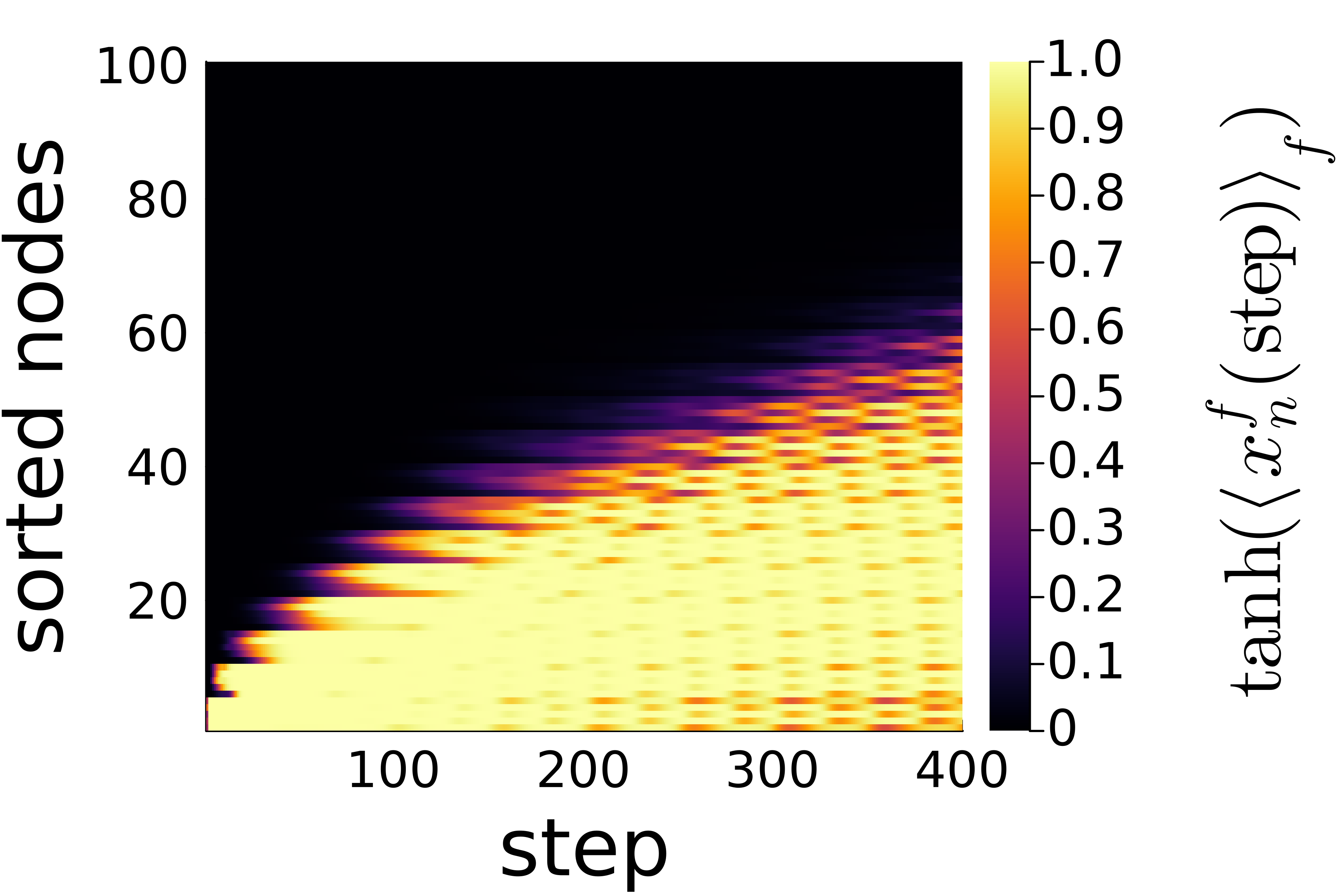}
    
    \includegraphics[width=0.3\columnwidth]{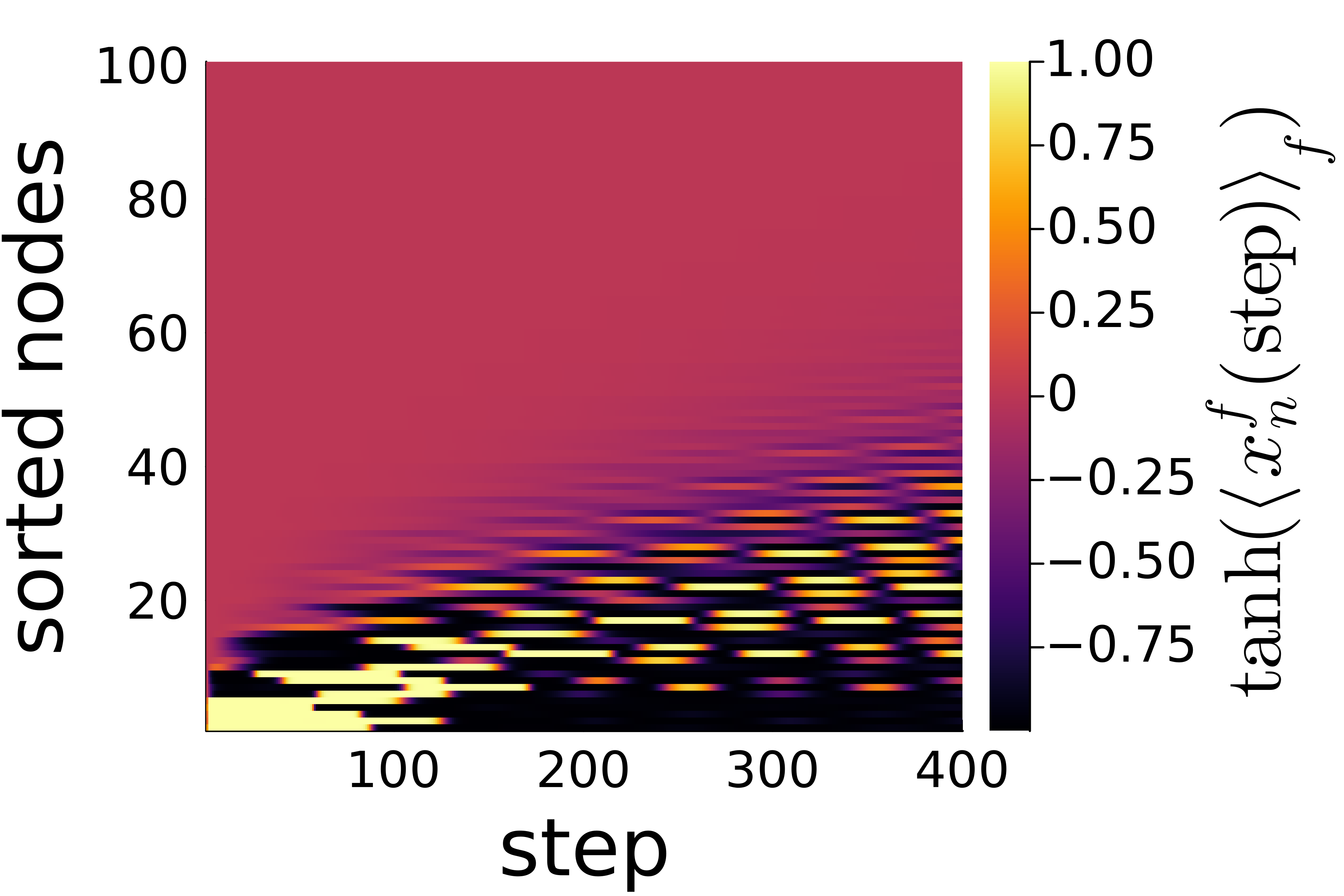}
    \includegraphics[width=0.3\columnwidth]{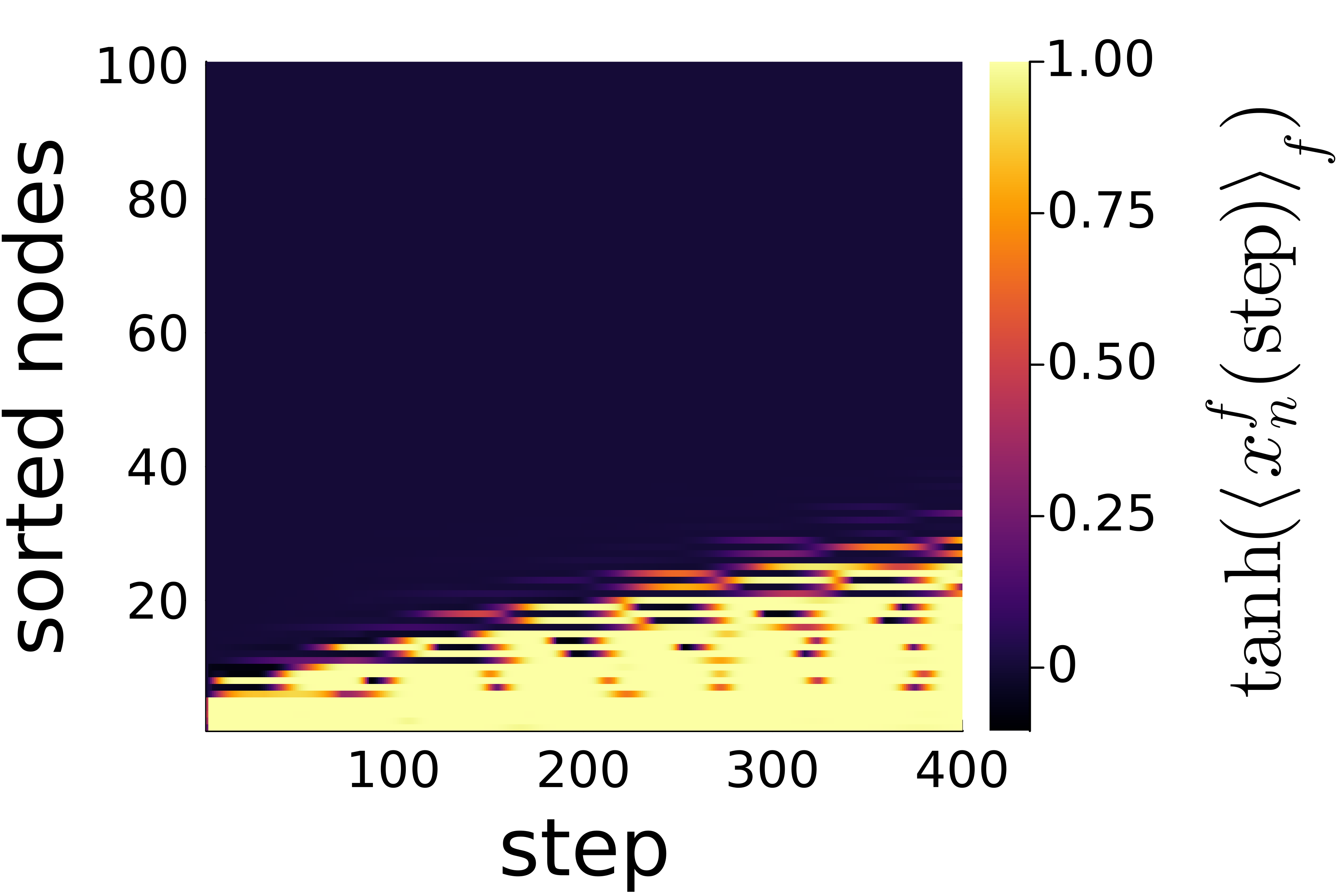}
    \includegraphics[width=0.3\columnwidth]{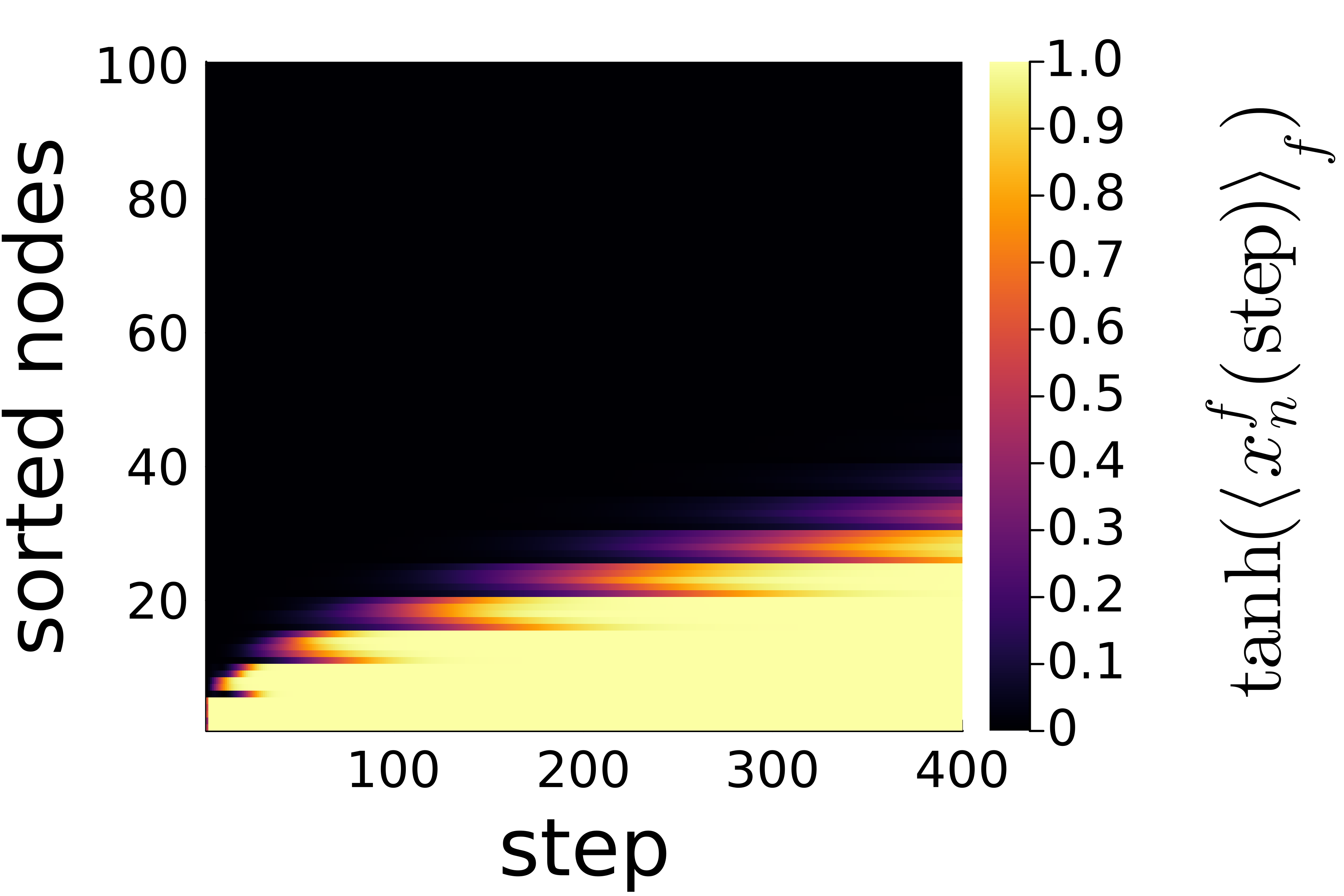}
    \caption{Non-oscillatory regime: Feature activation versus steps of the linear DB \Cref{eq:linDB} (top left), the non-linear DB 1-step layer \Cref{eq:linDB} + ReLU (top right), the linear MPNN layer \Cref{eq:MPNN} (bottom left), and the MPNN \Cref{eq:MPNN-sigma} without (middle) and with non-linear messages (bottom right). $d_n = d_f = 4$, same random weights.}
    \label{fig:spreading-non-oscil}
\end{figure}

\Cref{fig:spreading-non-oscil} provides example trajectories with random initial weights for which the MPNN and the DBGNN do not differ significantly. We suspect that random initialization leads to a washout of the directionality required to generate coherent propagation. How to robustly generate traveling activation remains an open research question.

\begin{figure}
    \centering
    \includegraphics[width=.5\columnwidth]{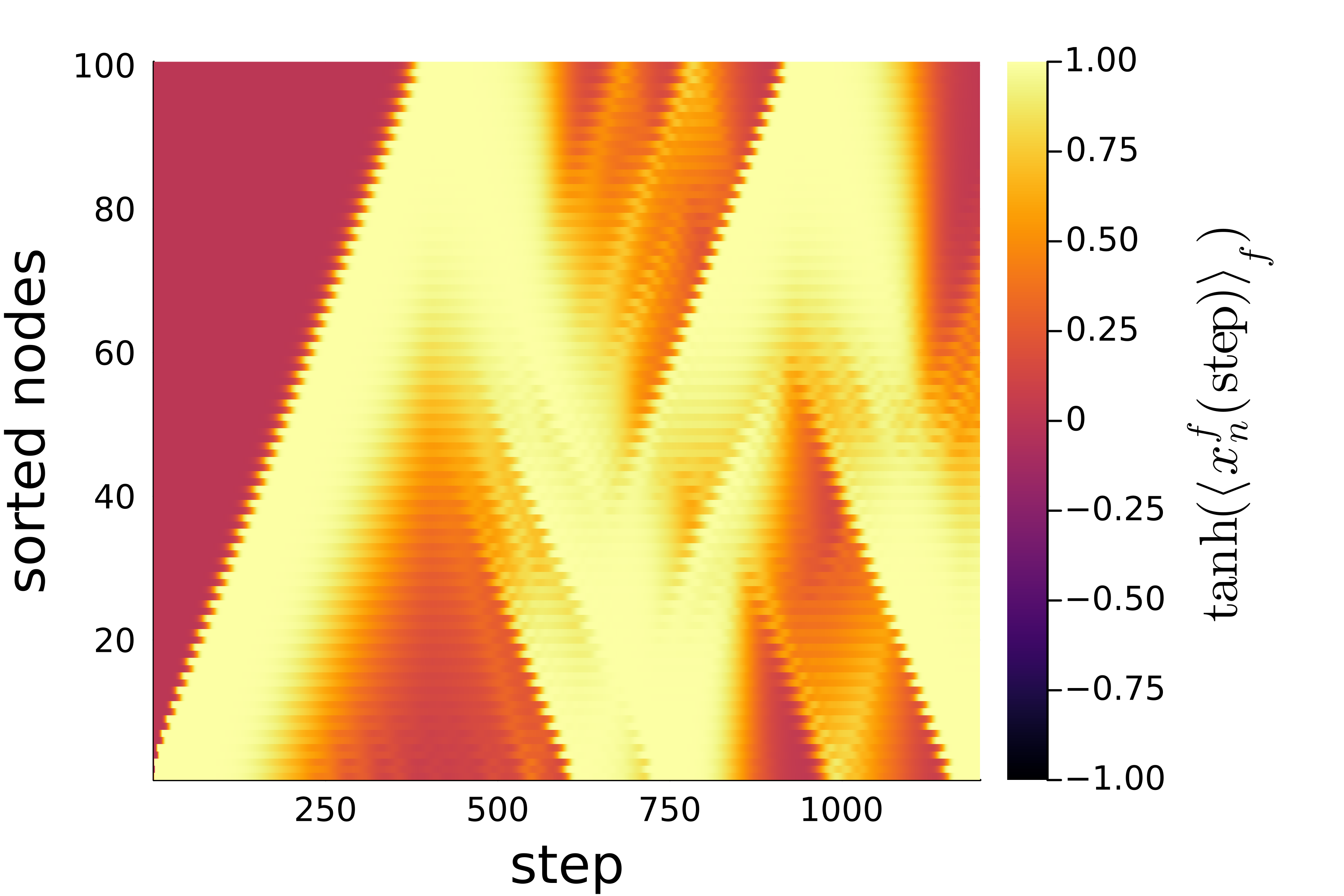}
    \caption{Oscillatory regime, full DB 1-step layer \Cref{eq:linDB} + non-linearity, ladder graph, $d_n = d_f = 4$.}
    \label{fig:ladder}
\end{figure}

We also provide two videos in the supplementary materials DB that show how a DBGNN layer can have activation that travels along a ladder graph and is reflected at the far edge compared to MPNN. The trajectory is provided in \Cref{fig:ladder}.

\section{Limitations}
The introduced DBGNN uses the Dirac operator, but no other special methods that could improve performance. It would be interesting to integrate other techniques, such as attention, to obtain better performance.

\section{Social impact}
We present a potentially powerful GNN for analyzing graph-structured data. We currently have no reason to believe that the new method introduces specific negative societal impacts beyond the general drawbacks of ML in general and improved graph neural networks in particular.

\end{document}